\documentclass[12pt]{article}
\usepackage{geometry}
\usepackage{times}
\usepackage{graphicx} % more modern
\usepackage{subfig}
\usepackage[numbers]{natbib}

\usepackage{algorithm}
\usepackage{algorithmic}

\usepackage{hyperref}

\usepackage{graphicx}
\usepackage{amssymb}
\usepackage{amsmath}
\usepackage{epstopdf}
\usepackage{multirow}
\usepackage{mathtools, bm}

\usepackage{color, multirow, dsfont}

\usepackage{url}

\usepackage{amsmath, amssymb, amsfonts}

\usepackage{graphicx}

\usepackage{mathtools, bm}

\usepackage{color, multirow, dsfont}

\newcommand{\bb}[1]{\mathbb{#1}}

\newcommand{\h}[1]{\hat{#1}}

\newcommand{\beq}{\begin{equation}}
\newcommand{\eeq}{\end{equation}}

\newcommand{\bal}{\begin{align}}
\newcommand{\eal}{\end{align}}

\newcommand{\ben}{\begin{enumerate}}
\newcommand{\een}{\end{enumerate}}

\newcommand{\bit}{\begin{itemize}}
\newcommand{\eit}{\end{itemize}}

\newcommand{\R}{\bb{R}}	% REALS
\newcommand{\norm}[1]{\left\lVert#1\right\rVert}
\newcommand{\algname}{HOMF}

\begin{document}

\title{Matrix Completion via Factorizing Polynomials}
\author{Vatsal Shah\thanks{vatsalshah1106@utexas.edu, The University of Texas at Austin}, Nikhil Rao\thanks{nikhilrao86@gmail.com, Amazon}, Weicong Ding \thanks{20008005dwc@gmail.com, Amazon} \footnote{This work was done when all three authors were affiliated with Technicolor Research Bay Area}
}
\maketitle

\begin{abstract} 
Predicting unobserved entries of a partially observed matrix has found wide applicability in several areas, such as recommender systems, computational biology, and computer vision. Many scalable methods with rigorous theoretical guarantees have been developed for algorithms where the matrix is factored into low-rank components, and embeddings are learned for the row and column entities. While there has been recent research on incorporating explicit side information in the low-rank matrix factorization setting, often implicit information can be gleaned from the data, via higher order interactions among entities. Such implicit information is especially useful in cases where the data is very sparse, as is often the case in real world datasets. In this paper, we design a method to learn embeddings in the context of recommendation systems, using the observation that higher powers of a graph transition probability matrix encode the probability that a random walker will hit that node in a given number of steps. We develop a coordinate descent algorithm to solve the resulting optimization, that makes explicit computation of the higher order powers of the matrix redundant, preserving sparsity and making computations efficient. Experiments on several datasets show that our method, that can use higher order information,  outperforms methods that only use explicitly available side information,  those that use only second order implicit information and in some cases, methods based on deep neural networks as well. 
\end{abstract}

\section{Introduction}
\label{sec:intro}
Matrix factorization into low rank components plays a vital role in many applications, such as recommender systems \cite{koren2009matrix}, computational biology \cite{biswas2015inferring} and computer vision \cite{haeffele2014structured}. 
Standard approaches to matrix factorization involves either explicitly looking for low-rank approximation via non-convex optimization \cite{recht2011hogwild,yu2012scalable} or resorting to convex, nuclear norm based methods \cite{candes2009exact,cai2010singular}.
%and solving it using semi-definite programs \cite{candes2009exact} or projected gradient methods that involve computing the SVD  \cite{cai2010singular}. 
%
These methods can be viewed as learning dense and low dimensional vector {\it embeddings} of the row and column entities (for example, users and movies), the inner products of which best approximate entries of the target matrix (for example, ratings). 
Such approaches for MF suffer from two drawbacks: practical datasets are often highly imbalanced, where some users might rate several items, whereas some users rarely provide feedback. Some of the items themselves are more popular than the others, making the observed entries highly nonuniform \cite{bhojanapalli2014tighter,mnih2008probabilistic}. Secondly, modern datasets also contain vast amounts of side information, incorporating which typically aids in prediction \cite{crandall2008feedback,zhou2012kernelized,jamali2009trustwalker,xu2013speedup,rao2015collaborative}. For example, one might have access to user social networks, product co-purchasing graphs, gene interaction networks and so on.

While it seems natural to enforce the learned embeddings to be faithful to the side information present, there also exists implicit information in the data. For instance, two users consistently providing same ratings to the same set of movies {\it implicitly} suggests the similarity in their preferences irrespective of the explicit links that may or may not exist between them. Indeed, low rank representations inherently look to achieve this by assuming that users (in this case) all lie in a common low dimensional subspace. However, explicitly modeling such an implicit similarity tends to yield better results. Recent advances have demonstrated that exploiting second-order co-occurrences of users and items indeed results in better recommendations \cite{Liang2016cofactor}. This implicit information is especially useful when the dataset is highly sparse and imbalanced, as is the case in most real world applications. 

In this paper, we develop a framework that can both \textit{seamlessly incorporate the provided side information if it exists, as well as use higher order information in the data to learn user and item embeddings}. The method we propose tackles both the issue of sparse data, as well as utilizing side information in an elegant manner. Furthermore, we develop an algorithm that precludes the computation of matrix higher powers\footnote{To be made clear in the sequel}, preserving the sparsity of the original problem and making computations highly efficient.

%
%In this paper, we develop a general framework, Higher Order Matrix Factorization (\algname), that can incorporate both implicit relationships via higher order information and explicit side information present in the data in a natural way to enhance prediction power. To clarify this further, our algorithm is not dependent on the presence of explicit side information, but if present we can efficiently assimilate it to improve our performance. Our method scales gracefully to incorporate information from higher order co-occurrences (beyond 2nd order). Also, we can efficiently incorporate pre-existing graph \footnote{note that graphs can be constructed via user or item features.} side information on both row and column entities of the matrix. The ease of scaling, utilization of side information coupled with its strong performance give our algorithm a clear advantage over most existing recommendation algorithms.
%

We model the target matrix \emph{combined} with side information as a  graph and then learn low-rank vector embeddings that best approximate the local neighborhood structure of the constructed graph. The key idea is that given a transition probability matrix for a graph $A$, the $i,j ^{th}$ entry of $A^k$ encodes the probability of jumping between nodes $i$ and $j$ in $k$ steps. Thus, even when the given data is very sparse, we can use it to glean missing entries, and use those to learn embeddings for users and items.

Let us consider a movie recommendation system. We construct a graph with all the users and movies as nodes. An (weighted, undirected) edge exists between a user node and movie node if the corresponding entry in the target matrix exists i.e. the user has rated the movie. Side information such as social networks amongst users, when present, are also used to form edges between user nodes. We aim to approximate the {\it multi-step transition probability}, i.e., probability of jumping between two nodes within a small, predefined number of hops, via a function proportional to the inner product between the node embeddings. This objective naturally encodes both the implicit and side information: transition is more likely between user nodes preferring a similar set of items; a dense clique in user social network also increase their probability of co-occurrence within a small number of steps. Noting that we consider co-occurrence within multiple steps, our embedding could account for higher order implicit similarities rather than the usual pair-wise similarity between users or movies, or just second order co-occurrences \cite{Liang2016cofactor}. Our experiments reinforce our claims that taking higher-order information indeed improves the prediction accuracy on a variety of datasets. We consider datasets with binary ratings as well as star-ratings, and those with and without side information. We call our method \algname : Higher Order Matrix Factorization.

\algname~is closely related to recent advances in representation learning in natural language processing  \cite{mikolov2013distributed,levy2014neural} and graphs \cite{perozzi2014deepwalk,tang2015line,tu2016max}. \cite{goyal2017graph} provides a comprehensive survey on various graph embedding techniques. While these works also consider co-occurrences of words within a small window in a sentence or nodes within a few steps on graphs, their objective is equivalent to factorizing the logarithm of such local co-occurrences \cite{yang2015comprehend}. 
% Similar to \algname, these approaches attempts to approximate functions of the co-occurrence pattern of words within a small local window in each sentence.   connected graph node embedding to word representation which is equivalent to  factorizing \cite{yang2015network}. 
In contrast, \algname~directly attempts to approximate the multi-step transition probabilities. Moreover, in this paper,  we explore various methods to {\it construct} edge weights of our graph from both rating-like and binary target matrix. 
Also, unlike the negative sampling and stochastic gradient descent approach in \cite{mikolov2013distributed,perozzi2014deepwalk}, we derive a coordinate descent based methods for solving the corresponding optimization problem. 
%
%address the computation bottleneck in calculating the multi-step transition probabilities and 
%
Finally, a recent line of work has focussed on using convolutional neural networks defined on graphs \cite{defferrard2016convolutional} to estimate the embeddings for users and items \cite{mgcnn}. The key idea in these methods is that node embeddings can be learnt via an aggregation of those of it's neighbors, and the ``scale" of the convolutional filter acts as the number of hops in a random walk. The method we propose is significantly simpler, and the optimization problem is more efficient. Despite this, we show that our method is competitive, and in several cases outperforms the methods that rely on deep neural nets to learn the embeddings.

%We verify that \algname~leads to better results on a number of different problem settings, with and without side information, feedback in standard star-rating or in binary form. 
%We showed that our algorithm leads to better results compared to the corresponding baselines in different settings. 

The rest of the paper is organized as follows: We first review related work. In the next section, we formally set up the problem we wish to solve and provide motivation for our algorithm. In Section \ref{sec:algo} we describe our algorithm, and comment on its computational complexity as well as remark on its generality. We provide extensive experimental results in Section \ref{sec:expt}. In Section \ref{sec:fw}, we discuss possible future directions and conclude the paper in Section \ref{sec:conc}. 
\subsection{Related Work}
\subsubsection{Incorporating Side Information:}
While several methods for vanilla matrix factorization (see \eqref{mf_standard}) have been proposed, incorporating explicit side information is a much more recent phenomenon. Given attributes or features for the row and/or column entities, \cite{xu2013speedup,jain2013provable} consider the row and column factors to be a linear combination of the features. This was extended to nonlinear combinations in \cite{gimc}. If graph information is known beforehand, then methods proposed in \cite{zhou2012kernelized,rao2015collaborative} can be applied to solve a slightly modified program: \eqref{mf_graph}. Further, note that one can construct the pairwise relationship between entities from the feature representations. Methods that can implicitly glean relationships between row and/or column entities have not been that forthcoming, an exception being \cite{Liang2016cofactor}, that uses second order information only. 

\subsubsection{Random Walks on Graphs:}
\cite{gori2007itemrank,xie2015edge} and references therein view collaborative filtering as functions of random walks on graphs. Indeed, the canonical user-item rating matrix can be seen as a bipartite graph, and several highly scalable methods have been proposed that take this viewpoint \cite{yun2014nomad}.  Methods that incorporate existing graph information in this context have also been studied \cite{golbeck2005computing,jamali2009trustwalker}.  \cite{fouss2007random} consider metrics such as average commute times on random walks to automatically figure out the similarity between nodes and apply it to recommender systems, but also note that such methods do not always yield the best results. Similarly, \cite{abbassi2007recommender} consider local random walks on a user-item graph, and resorts to a PageRank-style algorithm. Furthermore, they require the graph to have a high average degree, something most applications we consider will not have. In fact, canonical datasets might have a few nodes with high degree, but most nodes will have a very low degree. 

\subsubsection{Learning Node Embeddings on Graphs:}
Recently, ideas from learning vector representations of words \cite{mikolov2013distributed} have been used to obtain vector embeddings for nodes in a graph. Specifically, the authors in \cite{perozzi2014deepwalk} considered random walks on a graph as equivalent to ``sentences'' in a text corpus, and applied the standard Word2Vec method on the resulting dataset. \cite{yang2015network} showed that such a method is equivalent to factorizing a matrix polynomial after logarithmic transformations. In contrast, we directly aim to factorize a matrix polynomial with no transformations. Furthermore, we develop a method that makes computing the higher matrix powers in this polynomial redundant. These methods that rely on learning node embeddings of a graph can be viewed through a common lens: the embeddings are typically a weighted (nonlinear) combination of their neighbors in the graph. \cite{mgcnn} use the notion of graph convolutional neural networks to learn the embeddings, in the presence of side information presented as graphs. Finally, \cite{thyfriend}  consider a neighborhood-weighted scheme, and while the proposed method is not practical, provide theoretical justification for learning such embeddings. 
%The authors in \cite{GCN} apply the notion of graph convolutional neural networks, to learn node embeddings as a (nonlinear, weighted) combination of it's neighbors

\begin{figure*}[!htbp]
	\captionsetup{justification=centering}
	\centering
	\subfloat[][]{\includegraphics[width = 80mm, height = 50mm]{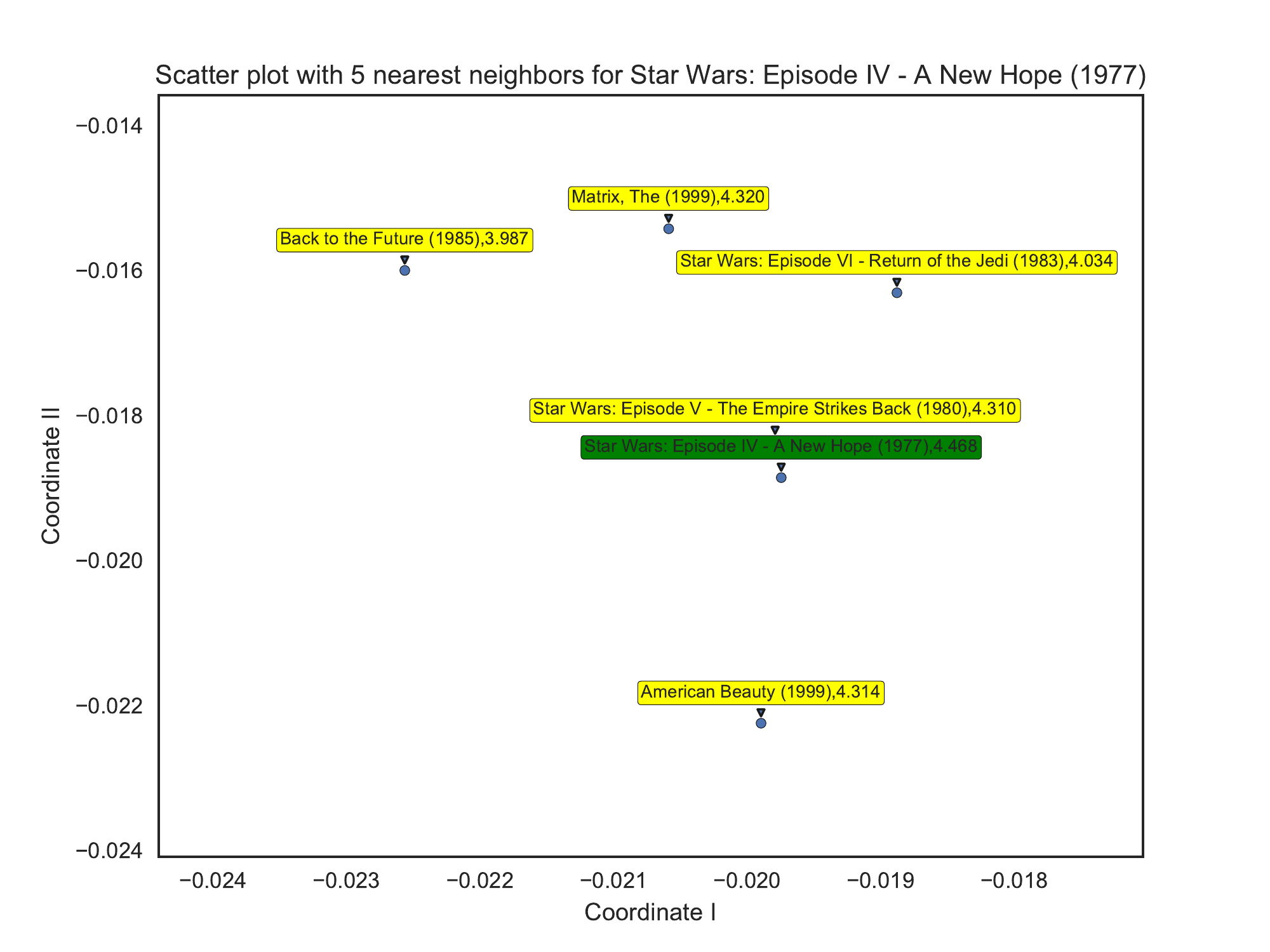}} \hfill
	\subfloat[][]{\includegraphics[width = 80mm, height =50mm]{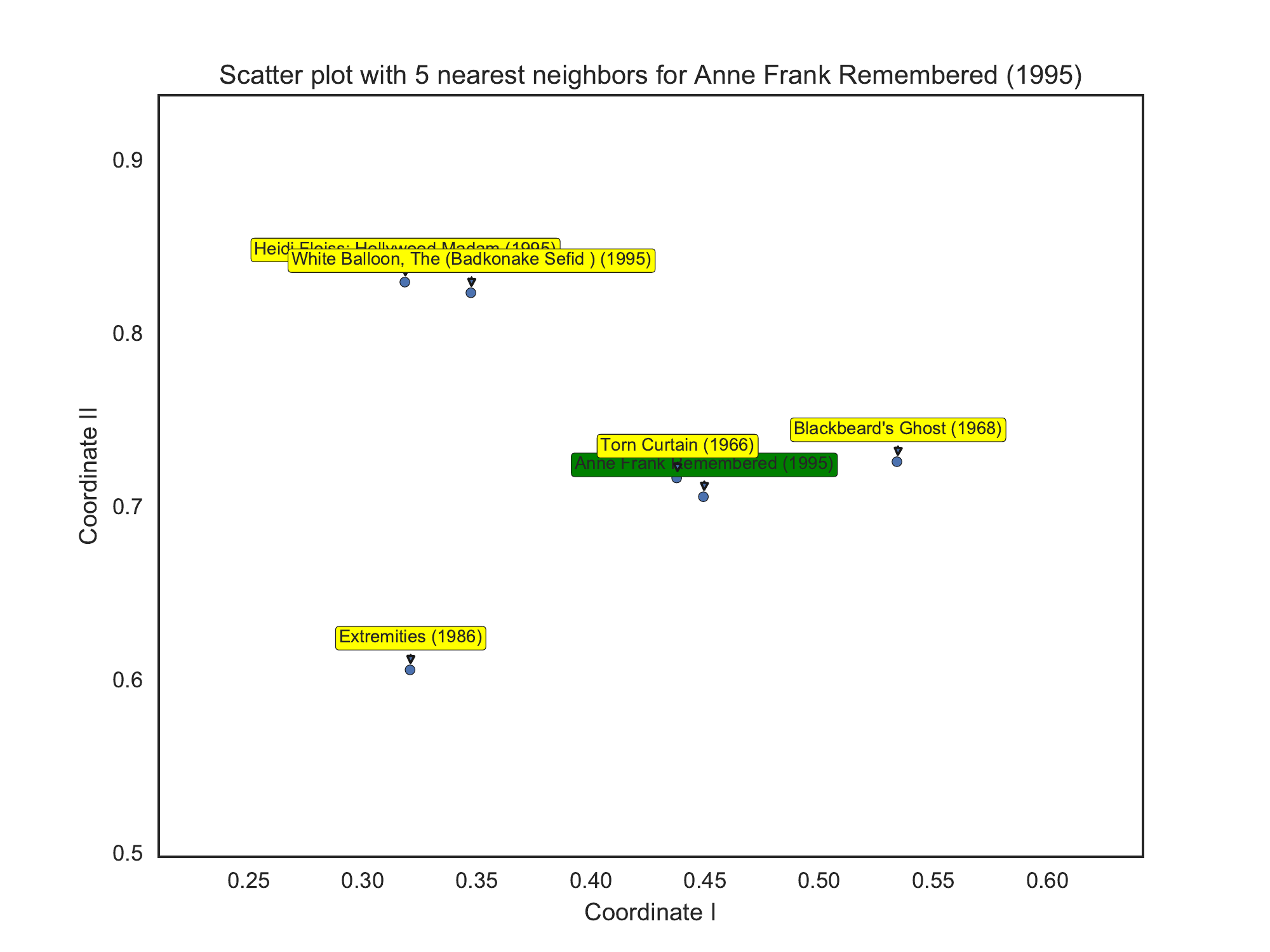}} \hfill
	\subfloat[][]{\includegraphics[width = 80mm, height =50mm]{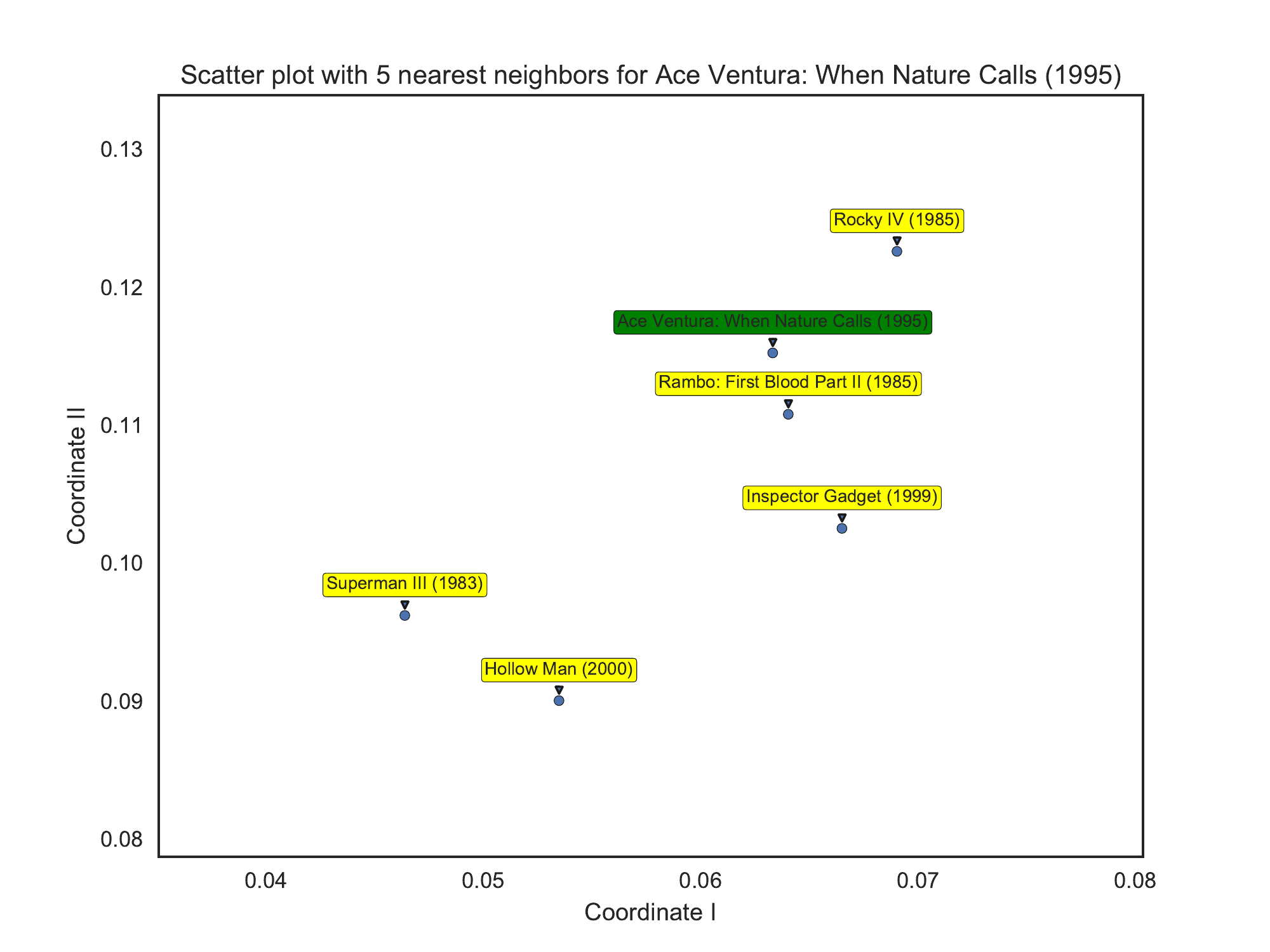}} \hfill 
	\subfloat[][]{\includegraphics[width = 80mm, height =50mm]{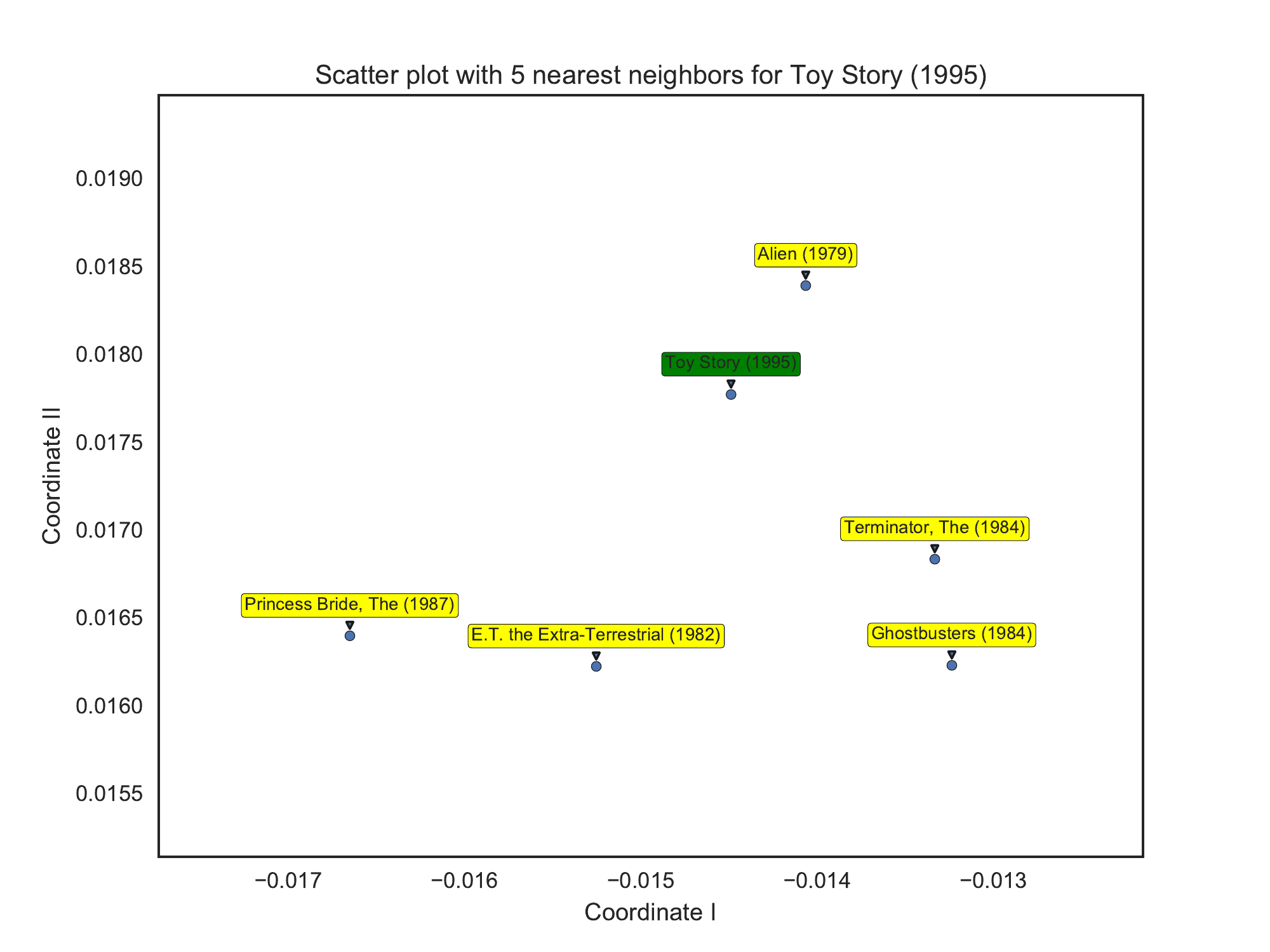}} \hfill \\
	\subfloat[][]{\includegraphics[width = 80mm, height =50mm]{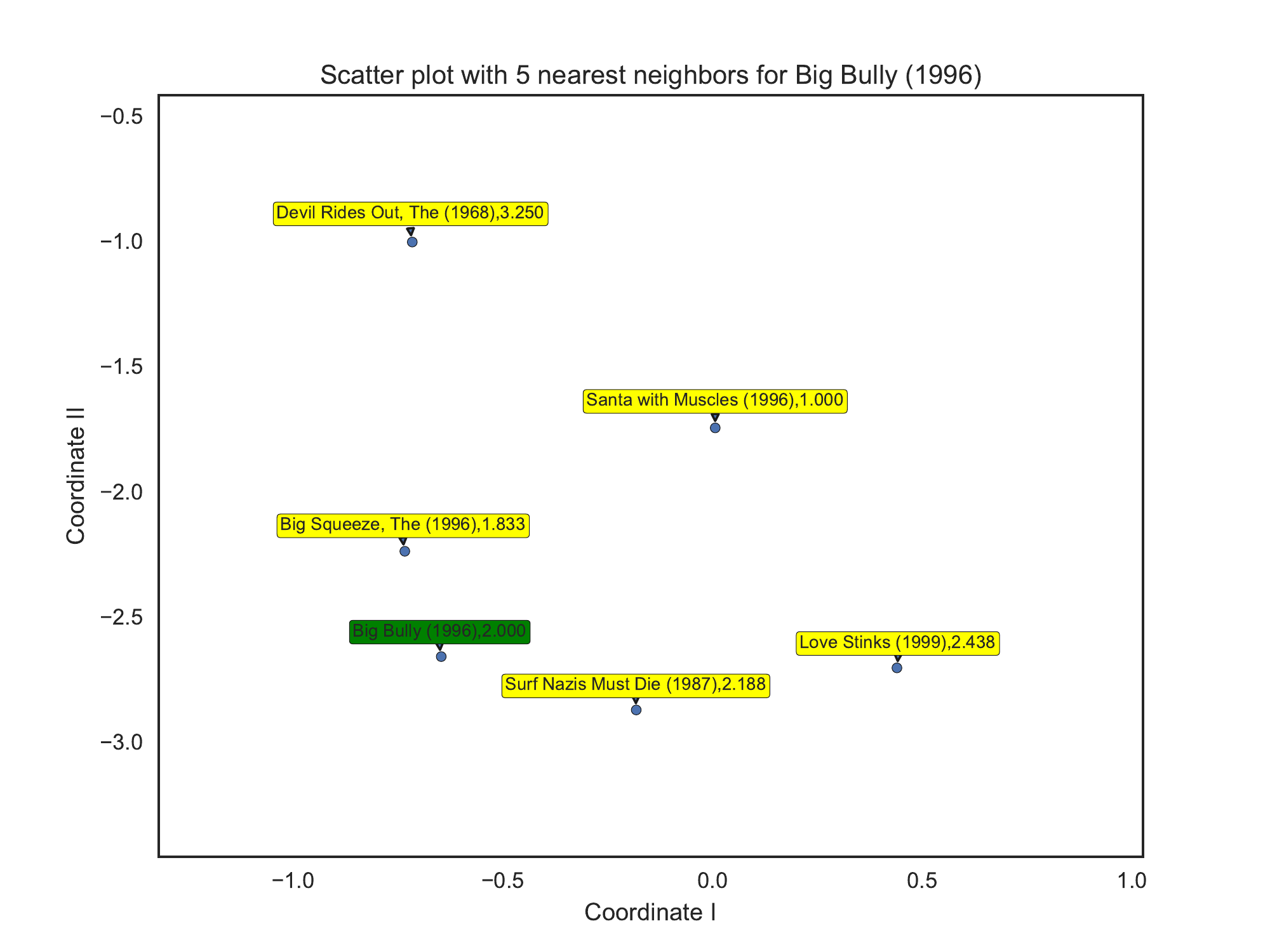}} \hfill 
	\subfloat[][]{\includegraphics[width = 80mm, height =50mm]{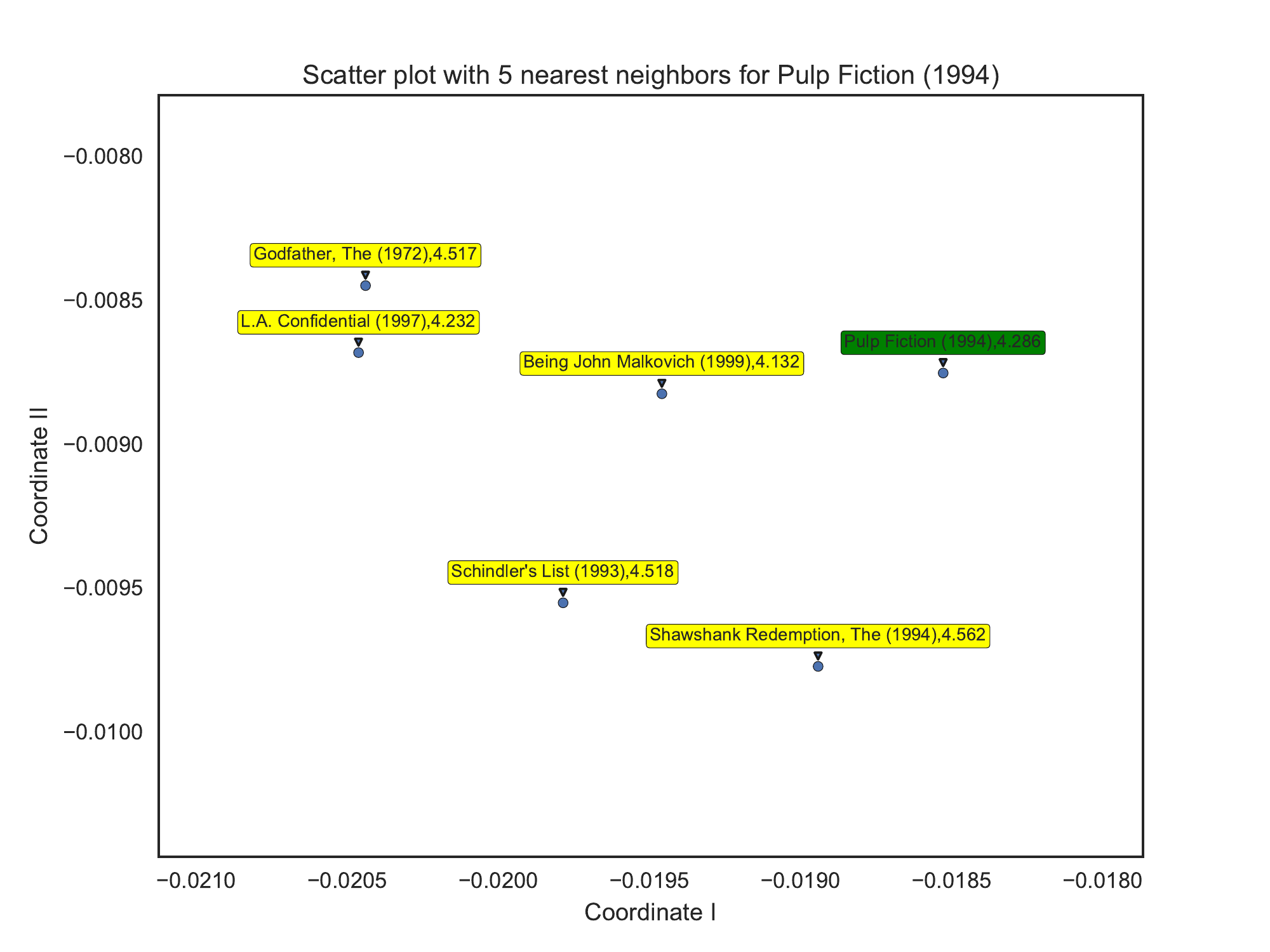}} \hfill 
	\caption{Five nearest neighbors (w.r.t L2 distance) for four different movies using vector embeddings generated from \algname. The movies are indicated with a green background and the nearest neighbor movies have a yellow background. Each label is a (movie, average rating in the training set) pair.}
	\label{fig:nn}
\end{figure*}

\section{Problem Setup} 
\label{sec:setup}
%
% some comments and edits made by WCD, Jan 06
%
% [title of this section? any better suggestion?]
%
We assume we are given a partially observed target matrix $R \in \R^{m \times n}$. Let the set of observed entries in $R$ be denoted by $\Omega$, where typically $| \Omega | \ll mn$. Given a rank parameter $k$, the goal in standard Matrix Factorization (MF) is to learn $k$-dimensional embedding vectors corresponding to the row and column indices of the matrix $R$. 
The standard matrix factorization algorithm aims to solve a problem of the form\footnote{bias variables are typically included, but we omit them from the text here for ease of explanation.}:
\begin{align}
\label{mf_standard}
\h{U}, \h{V} = \arg \min_{U,V} &\frac12 \| P_{\Omega}(R - UV^T) \|^2_F \notag\\
&+ \lambda \left( \| U \|^2_F + \| V \|_F^2 \right)
\end{align}
where $P_{\Omega}(\cdot)$ is the projection operator that retains those entries of the matrix that lie in $\Omega$, $U \in \R^{m \times k}$ and $V \in \R^{n \times k}$.

\noindent
{\bf Incorporating Side Information :} Following recent works by  \cite{zhou2012kernelized,rao2015collaborative}, it is possible to model the relationships between row (and column) entities via graphs. 
For example, in the case of a recommender system there might exist a graph $G_r$ among users, such as a social network,  and a product co-purchasing graph $G_c$ among items. 
Therefore, it is reasonable to encourage users belonging to the same social community or products often co-purchased together to have similar embeddings. 
Current state-of-the-art techniques proposed to solve the MF problem with side information encourage the embeddings of the row and column entities to be faithful with respect to the eigenspace of the corresponding Laplacians:
\begin{align}\label{mf_graph}
\h{U}, \h{V} = \arg \min_{U,V} &\frac12  \| P_{\Omega}(R - UV^T) \|^2_F \notag\\
&+\lambda \left( \mbox{Tr}(U^TL_rU) + \mbox{Tr}(V^TL_cV) \right)
\end{align}
where $L_r $ and $L_c$ are the graph Laplacians corresponding to $G_r$ and $G_c$ respectively. 

\noindent
{\bf Our approach- \algname:} While the aforementioned methods take side information into account, they cannot handle sparsity in the data. Specifically, if the observed matrix $R$ is highly sparse, and if the observed nonzero pattern is highly non-uniform, then it becomes hard to obtain reasonable solutions. In this paper, we look for a unified approach that can make use of $(1)$ explicit side information provided in the form of graphs and $(2)$ implicit information inherent in the data, via the use of higher order relationships among the variables (rows and columns). 

To this end, we propose constructing a (weighted, undirected) graph $\mathcal{G} = \{ \cal{V}, \cal{E} \}$ with $\cal{V}$ containing all the $m$ row entities and $n$ column entities. The edges in the graph are constructed as follows:
%$e_{ij} \in \cal{E}$ is a result of one of the following:
\vspace*{-1ex}
\begin{itemize}
	\item If two row entities $i,j$ are connected in $G_r$, we form an edge $e_{ij}\in\cal{E}$ with $e_{ij} = g_1(G_r(i,j))$. Here $g_1(.)$ is some non-negative, monotonic function of the edge weight in $G_r$. The same procedure is repeated for column entities that are connected in $G_c$.
	% If the similarity graphs $G_r$ and/or $G_c$ do not exist, $e_{ij}$ will be 0. 
	%
	
	\item If $i$ is a row node (e.g., user) and $j$ is a column node (e.g., a movie), we form an edge $e_{ij} = g_2(R(i,j))$ to encode the interactions observed in the data matrix $R$. $g_2(.)$ should also be a non-negative, monotonic function, and can potentially be the same as $g_1(.)$.

	\item We then scale the side-information edges with some weight parameter $\alpha\in (0,1)$. The data-matrix edges are scaled by $1-\alpha$. This allows us to trade-off the side information, and serves the same purpose as the parameter $\lambda$  in \eqref{mf_graph}.
\end{itemize} 

%We note that the mapping functions $g_1, g_2$ from data matrix and side information graph to edges in $\cal{G}$ are non-negative and monotonic. 
As a result of our graph construction, if a user rates a movie highly, then there will be a large edge weight between that user and movie. Similarly, if two movies are connected to each other via $G_c$, then again we expect the edge weight to be large. Figure \ref{fig:nn} fortifies these claims where we find the nearest neighbors for different movies using the vector embeddings generated by \algname. We see that movies rated similarly by users tend to cluster together, and if users' embeddings are close to each other, they will be recommended similar movies.

Typical choices for $g_1(.), g_2(.)$ include:
\begin{itemize}
	\item {\bfseries Exponential :} $g(x) = \exp(x)$
	\item {\bfseries Linear :} $g(x) = c \cdot x, ~\ $ for some $c \in \mathbb{R}$
	\item {\bfseries Step :} $g(x) = \begin{cases}  1 \hspace*{2cm}\text{ if }x>0\\  0 \hspace*{2.3cm}\text{ else} \end{cases}$
\end{itemize}

\noindent
Appropriately arranging the nodes, the overall adjacency matrix ${G}$ of our constructed graph can be represented as:
%\textcolor{red}{this needs to be explained better maybe?}
\begin{align}
\label{eq:mat}
{G} = \begin{bmatrix}
& \alpha g_1(G_r) ~~~ &(1 - \alpha)g_2(R)~~~~~~ \\ 
~~~ &(1 - \alpha) g_2(R)^T ~~~ &\alpha g_3(G_c)~~~~~ 
\end{bmatrix}
\end{align}
where with some abuse of notation, the functions $g_i(.)$ act element-wise on the arguments (i.e. matrices). We denote by $A$ the  \textit{row-normalized} version of this matrix ${G}$, so that each row of $A$ sums up to $1$. 
%where $\mbox{rnorm}[ \cdot ]$ is a function that normalizes the rows of the matrix, and  the $g_i()$ are functions that act element-wise on the arguments. The different subscripts merely indicate that the functions can be different monotonic functions, but they can all be the same. 
%
$A$ is thus a transition probability matrix (TPM), with $A_{ij}$ indicating the probability that a random walk starting at node $i$ jumps to node $j$ in one step. 

\noindent
Let
\begin{equation*}
\label{eq:TPMfunction}
f_T(A) :=  \dfrac{A+ A^{2}+\ldots+A^{T}}{T}
\end{equation*} 
for some positive integer $T$. Let $\Omega_T$ be the set of non-zero entries of $f_T(A)$. %A simple example of how to create $A$ from $R$ is presented in the appendix.

From a probabilistic point of view,  the $ij^{th}$ entry of $A^l$ is the probability of jumping from node $i$ to $j$ in $l$ random steps on $\cal{G}$ . Thus, $[f_T(A)]_{ij}$ is the probability of jumping from $i$ to $j$ {\it at least once} in $T$ steps in a random walk on $\cal{G}$. Hence, if this value is large, then it is safe to assume that nodes $i$ and $j$ are highly related to each other, even if there was no edge between them in $\cal{G}$. Our aim is to then ensure that the learned embeddings for these two nodes are similar. For example, two users rating the same set of movies have a higher probability of jumping between each other in a few steps on $\cal{G}$. In other words, if two users have rated movies in a similar fashion, then their future interactions will tend to be more alike than not. Similarly, two users with common friends in the side social network also have a higher transition probability to land on the other within a small number of jumps.

\noindent
Our objective in this paper  is to solve:
\begin{align}
\label{mf_homf}
\h{U}, \h{V} =  \arg \min_{U,V} &\frac12 \| P_{\Omega_T}(f_T(A) - UV^T) \|^2_F \notag\\
&+\lambda \left( \| U \|^2_F + \| V \|_F^2 \right)
\end{align}
where $U, V \in \mathcal{R}^{(m+n) \times k}$.  Letting $\h{u}_i$ be the $i^{th}$ row of $\h{U}$, we use $\h{u}_i\h{v}_j^T$ as a proxy for the predicted value of the corresponding entry in the target matrix. 
%More generally, $f_T(A)$ is able to capture higher order information
%

We note that the design parameter $\alpha$ controls the overall weights of the side information, and the number of steps $T$ determines how ``local" our search space in the graph will be. 
Overall,  \eqref{mf_homf} encourages nodes with higher co-occurrence probability within a small number of steps to have similar vector representations. The adverse effects of noisy/missing side information can be reduced by appropriately tuning $\alpha$. When there is no side information, the algorithm forces $\alpha$ to be 0 and ignores the side information matrix. Further motivation for factorizing $f_T(A)$ is provided in Section \ref{sec:fw}.

\subsection{Connections to Graph Convolutions}
Graph convolutions (e.g \cite{kipf2016semi}) have been recently developed as a generalization of spatial convolutions applied on graphs. The authors in \cite{defferrard2016convolutional} showed that graph convolutions can be efficiently computed by considering polynomials of the graph Laplacians, and performing matrix multiplications in the spectral domain. The key idea is that a node representation can be considered as a nonlinear combination of representations of it's neighbors, where the scale of the convolutional filter determines the number of hops from a node. Specifically, given a scale $T$, one can compute the coefficients $\theta_t$ of the polynomial filter
\[
g_\theta(\Lambda) := \sum_{t=0}^{T-1} \theta_t \Lambda^t
\] 
where $\Lambda$ is the singular values of the graph Laplacian. However, one first needs to compute the SVD of the Laplacian, which can be prohibitive even for moderately sized graphs. Furthermore, the iteration complexity in this case is equivalent to that of learning a Convolutional Neural Network. We work directly with the graph adjacency matrix and hence avoid SVD computations. Furthermore, in the next section we show that we need not even compute the higher powers of the adjacency matrix. We also do not learn the parameters of the linear combination $\theta_t$, and in fact including such a parametrization of the matrix $f_T(A)$ might yield further gains, at the expense of additional computations, something which we will leave for future work.  Despite not incorporating these complexities, we show that ~\algname performs comparably to and often outperforms Graph CNN based models.

\section{An Efficient Algorithm For \algname}
\label{sec:algo}
\begin{figure}[!h]
	\centering
	\includegraphics[height=45mm, width=60mm]{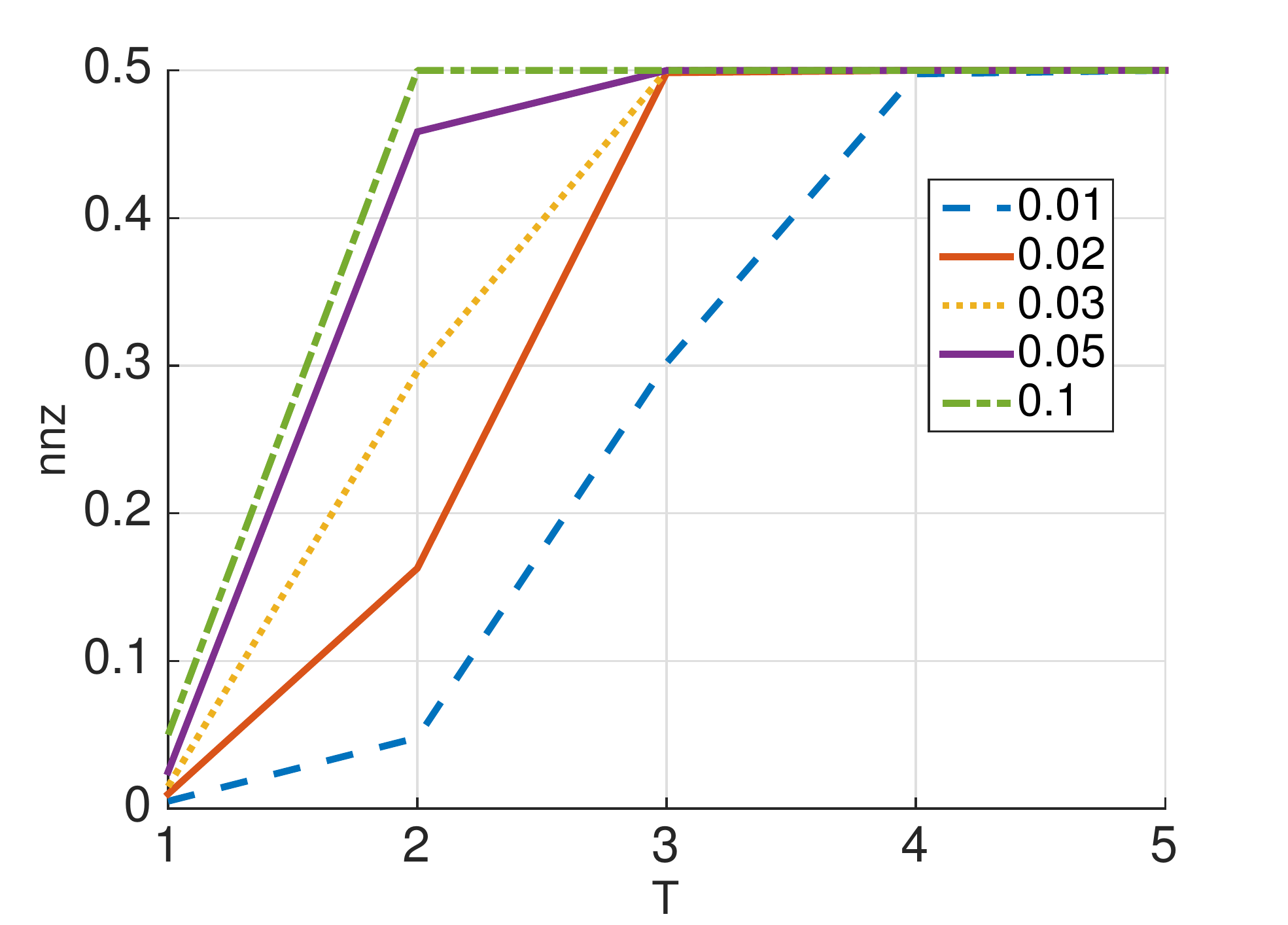}
	\vspace*{-2ex}
	\caption{Sparsity of $A^T$ for different power $T$ and sparsity level of observed set $\Omega$ for rating matrix $R$, where the numbers in the legend denote sparsity. The sparsity level is defined as the ratio of the size of $\Omega$ (i.e., the non-zero entries (nnz)) to the overall size of $R$. Note that even for $1\%$ sampling, the matrix saturates in terms of sparsity. (best seen in color)}
	\label{fig:sparsity}
\end{figure}

We now describe the algorithm to solve \eqref{mf_homf}. Note that once the matrix $f_T(A)$ is formed, the problem reduces to standard matrix factorization, for which highly efficient methods exist. However, obtaining $f_T(A)$ is expensive both from a computational and memory point of view. Indeed, regardless of the sparsity of $A$, $A^l$ for even small $l$, (say $l \geq 5$) will not be sparse.  Figure \ref{fig:sparsity} displays this phenomenon, where we created a random $1000 \times 1000$ matrix $R$, randomly select the observed set $\Omega$ with different sparsity levels, and constructed a block diagonal $A$ (eqn. \eqref{eq:mat}), without side information graphs ($G_r = G_c = 0$). \footnote{In this case, the maximal sparsity of $A^{l}, \forall l$ can be $0.5$}. Even for $1\%$ sparse $\Omega$, the multi-step transition $A^{T}$ quickly become dense. 

Given the size of modern datasets, storing such $A^l$ will be prohibitive, let alone storing $f_T(A)$. One thus needs an efficient method to solve the optimization problem we have. We will see in the sequel that we in fact need not compute $f_T(A)$, but only specific rows or columns of the matrix, which can be done efficiently.

We propose a coordinate descent method to alternatively update $U$ and $V$. Let $u_i^{(t)}$ (respectively $v_i^{(t)}$) be the $i^{th}$ row (respectively column) of $U$ (respectively $V$) at iteration $t$. We describe our method to update $V$ keeping $U$ fixed. The updates for $U$ with $V$ fixed will be analogous. At iteration $t$, the update equation for $V$ keeping the other entities fixed is as follows:
\begin{align}
\label{updateV}
%U^{(t)} &= \arg \min_V  \| P_{\Omega_T}(f_T(A) - U^{T}V^{(t-1)}) \|^2_F +\lambda \| U \|_F^2 \notag\\
V^{(t)} &= \arg \min_V  \| P_{\Omega_T}(f_T(A) - U^{(t-1)}V^T) \|^2_F +\lambda \| V \|_F^2 
\end{align}
% If all the entries of $f_T(A)$ are observed, this is a standard ridge regression problem with a closed form expression. It is, however, less straightforward to solve this without $f_T(A)$ being stored. 
This ridge regression problem can be solved column-wise without storing the $f_T(A)$ matrix. The key observation is that each column of $f_T(A)$ can be calculated efficiently via the following recursion:
\begin{align*}
f_T(A)\mathbf{e}_{i} =  \dfrac{1}{T} A(\mathbf{e}_{i}+ A(\mathbf{e}_{i}+A(\ldots )\ldots) )
\end{align*} 
where $\mathbf{e}_{i}$ is the $i^{th}$ canonical basis vector.  This computation requires only the storage of sparse matrix $A$ and can be executed on-the-fly by $T$ sparse matrix-vector multiplications. We outline the recursive details in Algorithm~\ref{alg:colsample}. An analogous procedure to the one mentioned in Algorithm~\ref{alg:colsample} can be applied to obtain the row wise updates.
\begin{algorithm}[!htbp]
	\caption{Recursive method for column sampling}
	\label{alg:colsample}
	\begin{algorithmic}
		\STATE {\bfseries Input:} matrix $A$, column  index $i$, steps $T$
		\STATE {\bfseries Initialize: } $a_1 = A_i, ~\ t = 1$
		\FOR{$t = 2,3,\cdots,T$}
		\STATE $a_{t} = a_{1} + Aa_{t-1}$
		\ENDFOR
		\STATE {\bfseries Return: } $f_T(A)_{\Omega^i_T} = \frac{1}{T} a_T$
	\end{algorithmic}
\end{algorithm}
% A key bottleneck of the procedure is sampling the $i^{th}$ column of $f_T(A)$. As mentioned above, computing and storing $f_T(A)$ is prohibitive, and near impossible on a single machine. We detail a method to sample a column of $f_T(A)$ (or rows) on-the-fly, requiring only the storage of $A$, and sparse matrix-vector multiplications in Algorithm \ref{alg:colsample}.
%
The updates for row-wise and column-wise optimization strategy lead us to solve the following problem:
\begin{align}
%u_i(t) = \arg \min_v \frac12 \| (f_T(A))_{\Omega^i_T} - uV \|^2_2 + \lambda \| u \|_2^2 \label{updateUcol}  \\ 
v_i(t) = \arg \min_v \frac12 \| (f_T(A))_{\Omega^i_T} - Uv \|^2_2 + \lambda \| v \|_2^2 \label{updateVcol}
\end{align}
where $\Omega^i_T$ is the set of nonzeros of the $i^{th}$ column of $f_T(A)$.
We note that Eq.~\eqref{updateVcol} can be solved using standard regularized least squares solvers, for example the conjugate gradient method as we adopted in this paper. Moreover, the updates for different columns can be parallelized. The empirical speedup via this parallelization is investigated later in the experiment section.  Algorithm \ref{alg:full_algo} details the pseudocode for solving \eqref{updateV}. A row-wise analogous version of Algorithm \ref{alg:full_algo} can be used to update $u_i(.)$.
\begin{algorithm}[!htbp]
	\caption{Procedure for solving \eqref{updateV}}
	\label{alg:full_algo}
	\begin{algorithmic}
		\STATE {\bfseries Input:} matrix $A$, steps $T$, fixed $U$, regularizer $\lambda$
		\FOR {$i = 1,2, \cdots, M$ {\bfseries in parallel }}
		\STATE Column sample $x := f_T(A))_{\Omega^i_T}$ using Algorithm \ref{alg:colsample}
		\STATE Solve \eqref{updateVcol} with $x, U, \lambda$ as inputs using conjugate gradients
		\STATE Set $v_i$ to be solution of above problem
		\ENDFOR
		\STATE {\bfseries Return: } $V$
	\end{algorithmic}
\end{algorithm}

\subsection{Computational Complexity}
\label{sec:complexity}
We briefly describe the computational complexity of our method. Each matrix-vector multiplication in Algorithm \ref{alg:colsample} can be performed in $O(nnz(A)(m+n))$ time. For a given $T$, the complexity for column sampling is then $O(nnz(A)(m+n)T)$. The main bottleneck in the conjugate gradients procedure to solve \eqref{updateVcol} is the Hessian-vector multiplication. The Hessian of \eqref{updateVcol} is $\left( U^TU + \lambda I \right) \in \R^{k \times k}$, and hence the complexity of the conjugate gradient method is $O(k^2(m+n))$. The per-iteration complexity of Algorithm \ref{alg:full_algo} is thus $O((nnz(A)T+k^2)(m+n))$, since the per-column update can be parallelized. Typically $k$ is a small number, so $k^2 \ll nnz(A)T$, and hence the complexity of the method is essentially linear in $m,n,T$ and the size of the data, including graph side information, since $nnz(A) = 2 nnz(R) + nnz(G_r) + nnz(G_c)$.

\subsection{Generality of \algname}
We see that \algname~ is highly general, and can efficiently incorporate pre-existing side information via the parameter $\alpha$. Indeed, the side information graphs can also yield interesting higher order interaction information, as we will see from our experiments. When side information is not present, we can obtain higher order information from only existing data. Furthermore, when $\alpha = 0$, $T=1$ effectively ignores higher order information, and is conceptually similar to the standard MF routine. $T=2$ in the same vein is similar to the co-factor method in \cite{Liang2016cofactor}. Finally, we can parameterize $f_T(A)$ as
\[
f_{T, \theta}(A) := \frac{1}{T} \sum_{i=1}^T \theta_i A^i
\]
and obtain further gains by learning the parameter vector $\theta \in \bb{R}^T$. We leave this for future work.

\section{Experiments}

%\vspace*{-3ex}

\begin{table*}[!t]
	\caption{Descriptions about the datasets used in our experiments}
	% Bayes on various data sets.}
	\label{tab:datasets}
	\begin{center}
		% dataset, type of data (binary, non binary), rows,  columns, nnz , graph , nnzgraph ,
		\begin{tabular}{| l | c | r | r | r | l | l | }
			\hline 
			\textbf{Dataset}	& \textbf{Type}	& $\mathbf{\#}$ \textbf{rows}	& $\#$\textbf{ columns}	& $\#$\textbf{ entries}	& \textbf{Graph present?}	& \textbf{$\#$ links} \\
			\hline
			ML-1M	& 0-5	& 6040		& 3952		& 1000208 	& no				& - \\ \hline
			% Epinion	& 0-5	& 5000		& 5000		& 179248  	& rows			& 487183 \\
			FilmTrust	& 0-5	& 1508 		& 2071		& 35497		& rows			& 1632 \\ \hline
			Gene	& Binary	& 1071		& 1150		& 2908		& rows and columns	& 7424 (rows) \\
						& 		& 			& 			& 			& 				& 437239 (columns)	\\	
			\hline
		\end{tabular}
	\end{center}
\end{table*} 

\label{sec:expt}
In this section, we test our method against standard matrix factorization and other state-of-the-art methods that use graph side information. We have attempted to use a variety of datasets, some with additional graph containing side information, and some without to demonstrate the universality of our algorithm. We also consider a binary dataset with graphs. Details about the datasets are provided in Table \ref{tab:datasets}.  MovieLens 1 million (ML-1M) is a standard movie recommendation dataset  \footnote{\url{https://grouplens.org/datasets/movielens/1m/}}
\cite{harper2016movielens}. The FilmTrust dataset \cite{guo2013novel} has a similar task as that of ML1m, but also has user-user network given \footnote{\url{http://www.librec.net/datasets.html}}. 
The Gene dataset is a binary matrix, where the task is to determine what genes are useful for predicting the occurrence of various diseases \footnote{\url{http://bigdata.ices.utexas.edu/project/gene-disease/}}. This dataset also contains both gene-gene interaction network data and a disease co-morbidity graph.  
Our primary motivation between choosing such \textit{diverse datasets} is to emphasise the \textit{effectiveness and flexibility} of our algorithm in handling different types of data, side information without significant additional preprocessing.

For ML-1M and FilmTrust, we generate a random 80/20 Train-Test split of the data. For Gene, we use the split provided online. Since this is a one-class classification problem, we randomly sampled the same number of negative samples as in the training data, to act as the negative class for classification. 

%For the MovieLens 1 million (ML1m) and FilmTrust datasets, we create a random 80/20 train/test split of the target data. For the Gene dataset we create negative samples we use the train/test splits provided in \url{http://bigdata.ices.utexas.edu/project/gene-disease/}.
%For the Epinions data, we retain the top 5000 users with most ratings provided, and intersect it with the top 5000 movies that have been rated the most. 

\subsection{Baselines, Parameters and Initialization}
We fixed the rank (k) for all methods to be 10. For comparing against standard matrix completion and the graph based counterparts (equations \eqref{mf_standard} and \eqref{mf_graph}), we used the GRALS method \cite{rao2015collaborative}\footnote{Evaluated using the code available online: https://www.cs.utexas.edu/~rofuyu/exp-codes/grmf-exp.zip}. We also compared to Weighted MF \cite{hu2008collaborative}\footnote{Evaluated using the code available online: https://github.com/benanne/wmf}.
For FilmTrust dataset with side information graph, we also compared against the TrustWalker algorithm \cite{jamali2009trustwalker} which makes recommendations using random walks on graphs. We also compare against CoFactor \cite{Liang2016cofactor}, which can be seen as a specialization of our method to use only second order information, and the graph convolutional method introduced recently in \cite{mgcnn} (MGCNN), where the authors show improved performance over GRALS for the same tasks. 
%As a means to test against a method that uses second order information, we also used the CoFactor method \cite{Liang2016cofactor}
%

We implemented our method in python using the multiprocessing framework to parallelize the method in Algorithm~\ref{alg:full_algo}~\footnote{Python Implementation of our algorithm is publicly available at https://github.com/vatsal2020/HOMF }.  Wherever applicable, we varied $\lambda \in \{10^{-4}, 10^{-3}, \ldots, 10^2 \}$ and cross-validate $\lambda$ on a fixed validation set (20$\%$ of the training data). When the graph side information is available, we also varied $\alpha \in \{ 0.15, 0.25, 0.5, 0.75 \}$.
% We plan to make our code available online for research purposes. 
We consider random walks of lengths $\{2,4,6, \dots T_{max}\}$ for all datasets. For ML-1M, FilmTrust and Gene, $T_{max}$ was fixed at $8, ~ 6$ and $6$ respectively. We initialized the factor matrices $U, V$ such that each entry is an independent uniform $[0,1]$ random variable. For the method in \cite{mgcnn}, we used the default neural network parameters, set the rank = 10 to be consistent with our models, and used the output of the graph CNN as the user and movie embeddings. The default ``bandwidth" for the graph CNNs is set to $5$, which we refer to as (MGCNN-def). We also varied this bandwidth using the same $T$ as in our method, which we call MGCNN in the results below.

%
%%%%%%%%%%%%%%%%%%%%%%%%%%%%%%%%%%%%%%%%%%%%%%%%%%
%%%%%%%%%%%%%%%%%%%%%%%%%%%%%%%%%%%%%%%%%%%%%%%%%%
%%%%%%%%%%%%%%%%%%%%%%%%%%%%%%%%%%%%%%%%%%%%%%%%%%
%%%%%%%%%%%%%%%%%%%%%%%%%%%%%%%%%%%%%%%%%%%%%%%%%%
%%%%%%%%%%%%%%%%%%%%%%%%%%%%%%%%%%%%%%%%%%%%%%%%%%
%%%%%%%%%%%%%%%%%%%%%%%%%%%%%%%%%%%%%%%%%%%%%%%%%%
%%%%%%%%%%%%%%%%%%%%%%%%%%%%%%%%%%%%%%%%%%%%%%%%%%
%%%%%%%%%%%%%%%%%%%%%%%%%%%%%%%%%%%%%%%%%%%%%%%%%%
%%%%%%%%%%%%%%%%%%%%%%%%%%%%%%%%%%%%%%%%%%%%%%%%%%
%%%%%%%%%%%%%%%%%%%%%%%%%%%%%%%%%%%%%%%%%%%%%%%%%%
%

\subsection{Evaluation Metrics}
For the Gene dataset with binary observations, we computed the standard AUC score on the  test set. 
For non-binary data (ML-1M and FilmTrust), we compute Precision, Recall, Mean Average Precision (MAP), and Normalized Discounted Cumulative Gain (NDCG), all $@$ various thresholds (K). Note that since we are factorizing a nonlinear transformation of the ratings matrix $R$, the RMSE will not be a useful metric to compare. For the ML-1M dataset, we determine that a movie rated by a user is a true positive if the corresponding rating is 5 while $K=\{5,10\}$. For the FilmTrust dataset, we determine that a movie is a true positive if the corresponding rating is at least 3 and corresponding values of $K$ are set to $1$ and $2$ since the dataset is small and it is hard to find many highly rated movies for most users. The \textit{K} and threshold values are determind by the average number of highly rated movies for each user which is consistent with most available literaute. For the sake of completeness and to prevent confusion, we provide explicit formulas for the metrics used. Let $i_1, \ldots, i_{n_u}$ be items in test set rated by user $u$ {\it sorted} by the predicted score. Let $\mathcal{I}(u,i)=1$ if user $u$ rated item $i$ as relevant in the ground-truth test data and $0$ otherwise. For simplicity, let $\mathcal{I}_u = \sum_{\ell=1}^{n_{u}} \mathcal{I}(u, i_{\ell})$ be the total number of relevant items per user. We first calculate for each user $u$:
\vspace*{-0.25cm}
\begin{align*} 
\text{Precision}@(K,u)& =\sum_{j=1}^{K} \mathcal{I}(u, i_{j})/K\\
~~~\text{Recall}@(K,u) &=\sum_{j=1}^{K} \mathcal{I}(u, i_{j})/\mathcal{I}_u
\end{align*}
\begin{align*}
\text{AP}@(K,u)& =\sum_{j=1}^{K} \text{Precision}@(j,u)/\min(\mathcal{I}_u, K) 
\end{align*} 
and then average across all the users to get Precision@K, Recall@K, and MAP@K. For NDCG, we first calculate

\begin{align*}
\text{DCG}@(K,u)& =\sum_{j=1}^{K} \frac{2^{\mathcal{I}(u,i_{j})}-1}{\log(i+1)}
\end{align*}
and $\text{IDCG}@(K,u)$ based on the ordered ground truth ranking of all the items. We obtain $\text{NDCG}@(K,u) = \text{DCG}@(K,u)/\text{IDCG}@(K,u)$ and average it across all users.

\begin{table*}[t!]
	\captionsetup{justification=centering}
	\caption{Comparison of various algorithms using Top-N evaluation metrics. \algname~ performs better than the methods compared consistently. Bold values indicate the best result among the methods considered. The exponential activation function was used to obtain the following results, and we set $g_1(.) = g_2(.)$ for our graphs. }
	\label{tab:results}
	\begin{center}
		\begin{tabular}{| l | l | c | c | c | c | c | c | c | c |}
			\hline 
			% \abovespace\belowspace
			{\textbf{Data}} &{\textbf{Algos}}  &\multicolumn{2}{|c|}{\textbf{Precision}} &\multicolumn{2}{|c|}{\textbf{Recall}} &\multicolumn{2}{|c|}{\textbf{MAP}} &\multicolumn{2}{|c|}{\textbf{NDCG}}\\
			\hline
			& & $@5$ & $@10$ & $@5$ & $@10$ & $@5$ & $@10$ & $@5$ & $@10$ \\
			\hline
			ML-1M 	& MF 	 	& 0.316	& 0.311	& \textbf{0.598}	& 0.614	& 0.462	& 0.474	& 0.692	& 0.707	\\
			&Cofactor &0.360 &0.320 &0.530 & 0.641	& 0.480	& 0.514	& 0.492	& 0.531	\\
			&WMF  & 0.357 & 0.329 &0.532 & 0.649 &0.475 & \textbf{0.535} &0.653 & 0.744 \\
			& MGCNN-def 	 	& 0.367	& 0.323	& 0.549	& 0.649	& 0.496	& 0.530	& 0.740	& 0.744	\\
			& MGCNN 	 	&0.367	&0.329  & 0.578	 & \textbf{0.665}	& 0.496	& 0.533	& 0.740	& 0.744	\\
			&\algname	& \textbf{0.370}	& \textbf{0.331}	& 0.544	& 0.662	& \textbf{0.499}	& 0.529	& \textbf{0.744}	& \textbf{0.749}	\\		
			%\hline
			%Epinions	& MF 	 	& \textbf{0.436}	& \textbf{0.395}	& 0.812	& \textbf{0.859}	& \textbf{0.742}	& \textbf{0.735}	&\textbf{ 0.816}	& \textbf{0.802}	\\
			%		& GRALS 	 	& 0.434	& 0.388	& 0.801	& 0.846	& 0.720	& 0.727	& 0.796	& 0.799	\\
			%       & TrustWalker & 0.412  & 0.372 & 0.756  & 0.772 & 0.684 & 0.706 & 0.765  & 0.745\\
			% 		&\algname	& 0.426	& 0.390	& \textbf{0.816}	& 0.830	& 0.710	& 0.730	& 0.789	& 0.775	\\ 	
			\hline
			& & $@1$ & $@2$ & $@1$ & $@2$ & $@1$ & $@2$ & $@1$ & $@2$ \\
			%  \hline
			%Epinions	& MF 	 	& \textbf{0.499}	& \textbf{0.485}	& 0.373	& 0.644	& \textbf{0.640}	& \textbf{0.667}	&\textbf{ 0.769}	& \textbf{0.812}	\\
			%		%& GRALS 	 	& 0.434	& 0.388	& 0.801	& 0.846	& 0.720	& 0.727	& 0.796	& 0.799	\\
			%       & TrustWalker & 0.442  & 0.437 & 0.321  & 0.575 & 0.567 & 0.589 & 0.719  & 0.758\\
			% 		&\algname	& 0.483	& 0.466	& \textbf{0.394}	& \textbf{0.658}	& 0.628	& 0.660	& 0.763	& 0.807	\\ 		
			%\hline
			% & & $@1$ & $@2$ & $@1$ & $@2$ & $@1$ & $@2$ & $@1$ & $@2$ \\
			\hline
			FilmTrust & MF 	 	& 0.701	& 0.633	& 0.345	& 0.436	& 0.795	& 0.744	& 0.761	& 0.747	\\
			& TrustWalker	& 0.506	& 0.497	& 0.316	& 0.456	& 0.598	& 0.607	& 0.584	& 0.568	\\
			& GRALS 	 	& 0.752	& 0.740	& 0.365	& 0.492	& 0.812	& 0.801	& 0.772	& 0.770	\\
			&WMF          & 0.750 & 0.746 & 0.366 & 0.501 & 0.803 & 0.761 & 0.775 & 0.771 \\ 
			& Cofactor 	 	& 0.716	& 0.706	& 0.349	& 0.469	& 0.773	& 0.762	& 0.716	& 0.732	\\
			& MGCNN-def 	 	& 0.751	& 0.741	& 0.370	& 0.495	& 0.805	& 0.793	& 0.772	& 0.767	\\
			& MGCNN 	 	& \textbf{0.761}	& \textbf{0.748}	& 0.367	& 0.499	& \textbf{0.822}	& 0.793	& 0.777	& \textbf{0.775}	\\
			&\algname	& 0.754	& 0.745	& \textbf{0.375}	& \textbf{0.502}	& 0.816	& \textbf{0.802}	& \textbf{0.778}	& 0.773	\\			
			\hline	
		\end{tabular}
	\end{center}
\end{table*}

\subsection{Results}
Table \ref{tab:results} summarizes the results obtained on the test set in ML-1M and FilmTrust. 
%Results for all parameters considered are provided in the Appendix {\color{red} Are we doing this? }. 
We see that \algname consistently outperforms standard matrix factorization, and also the version that uses graph side information (GRALS) as well as CoFactor.  In some cases, the performance gap is significant. MGCNN is a much more complex model, and it is often better than the other methods as the authors in \cite{mgcnn} show. We see that \algname~ is for the most part competitive with, and often outperforms MGCNN, despite being significantly simpler to optimize and having fewer parameters to learn. 
%Another important point to note is that \algname~is not universally better, and indeed in some settings (eg. Epinions) the standard MF method seems to yield the best performance. This suggests that the choice of method to use largely depends on the dataset, and there is no universal silver bullet for such problems. 

\begin{table}[!bp]
	\captionsetup{justification=centering}
	\begin{center}
		\begin{tabular}{| l | l | c |}
			\hline
			\textbf{Data}	& \textbf{Method}	& \textbf{AUC} \\
			\hline
			Gene		& MF		& 0.546 \\
			& GRALS		& 0.572 \\
			& Cofactor     & 0.506 \\
			& WMF & 0.568 \\
			& MGCNN-def     & 0.569 \\
			& MGCNN     & 0.599 \\
			& \algname (exp)	& 0.623 \\
			& \algname (linear)	& \textbf{0.630} \\
			\hline
		\end{tabular}
	\end{center}
	\caption{Comparison of various algorithms for Gene dataset}
	\label{tab:gene}
\end{table}

In Table \ref{tab:gene}, we show the results obtained on the Gene-Disease dataset. In this setting, with rich graph information encoding known relationships between the entities, \algname~ significantly outperforms the competing methods. This suggests that there are hidden, higher order interactions present in the data, which the given graphs do not fully capture by themselves. This opens an interesting avenue for further research, especially in domains such as computational biology where obtaining data is hard, and hence hidden information in the data is even more valuable. Furthermore, a simpler parameterization helps in this case, since the data is sparse and NN based methods typically need a lot of data to train.

\subsection{Effect of $T$, $g_i(.)$ and $\alpha$}

\begin{figure*}[!ht]
	\captionsetup{justification=centering}
	\caption{Performance of HOMF on the ML-1M (letf 2) and FilmTrust (right 2) datasets, as T and $\alpha$ is varied. The Y axis for (a) and (c) is precision, and for (b) and (d) is recall. We also see that sometimes the $\exp()$ function is better and sometimes it is worse than linear. (best seen in color)}
	\centering
	\subfloat[][]{  \includegraphics[width = 35mm, height = 30mm]{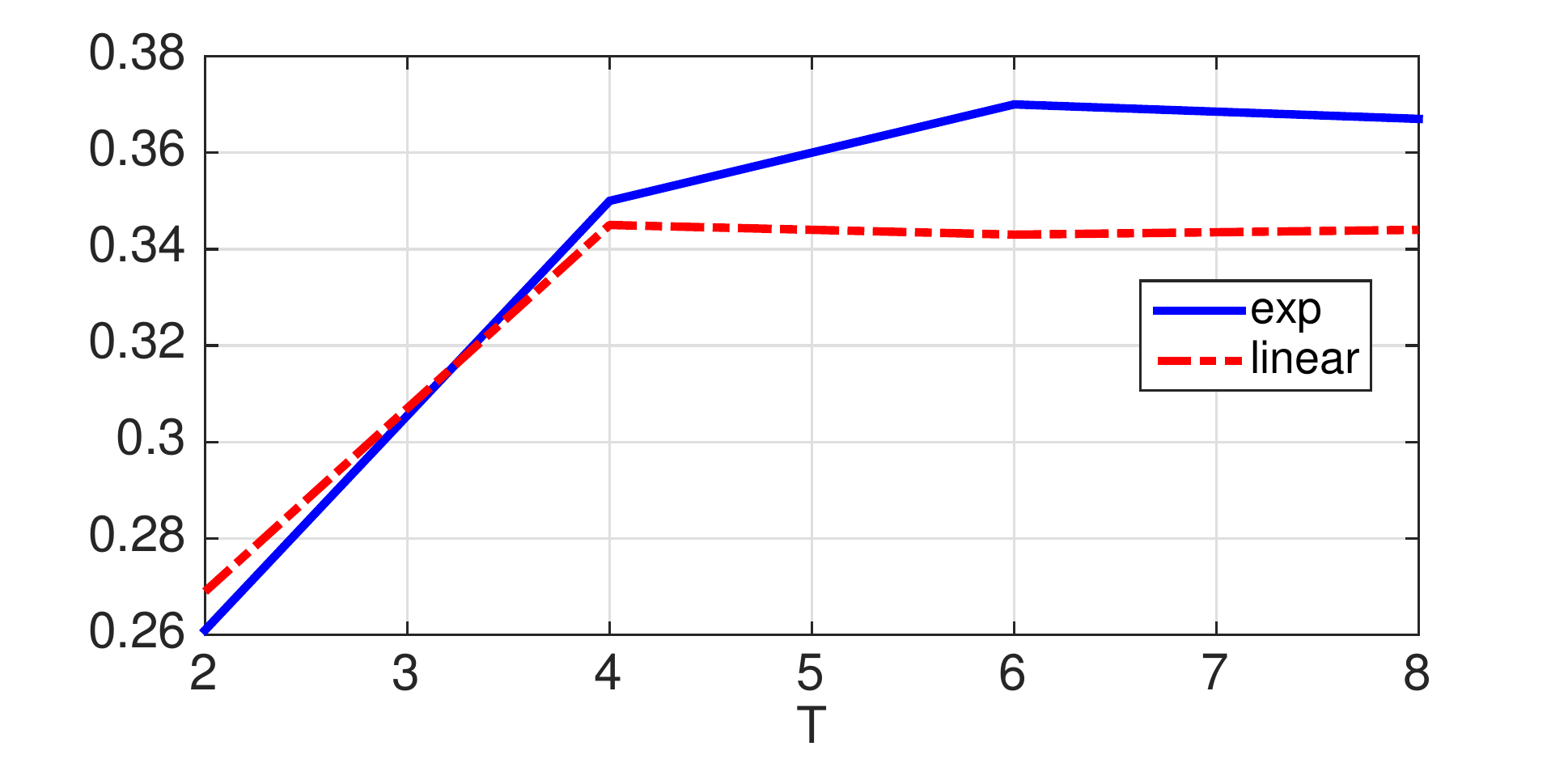}} \hfill
	\subfloat[][]{  \includegraphics[width = 35mm, height = 30mm]{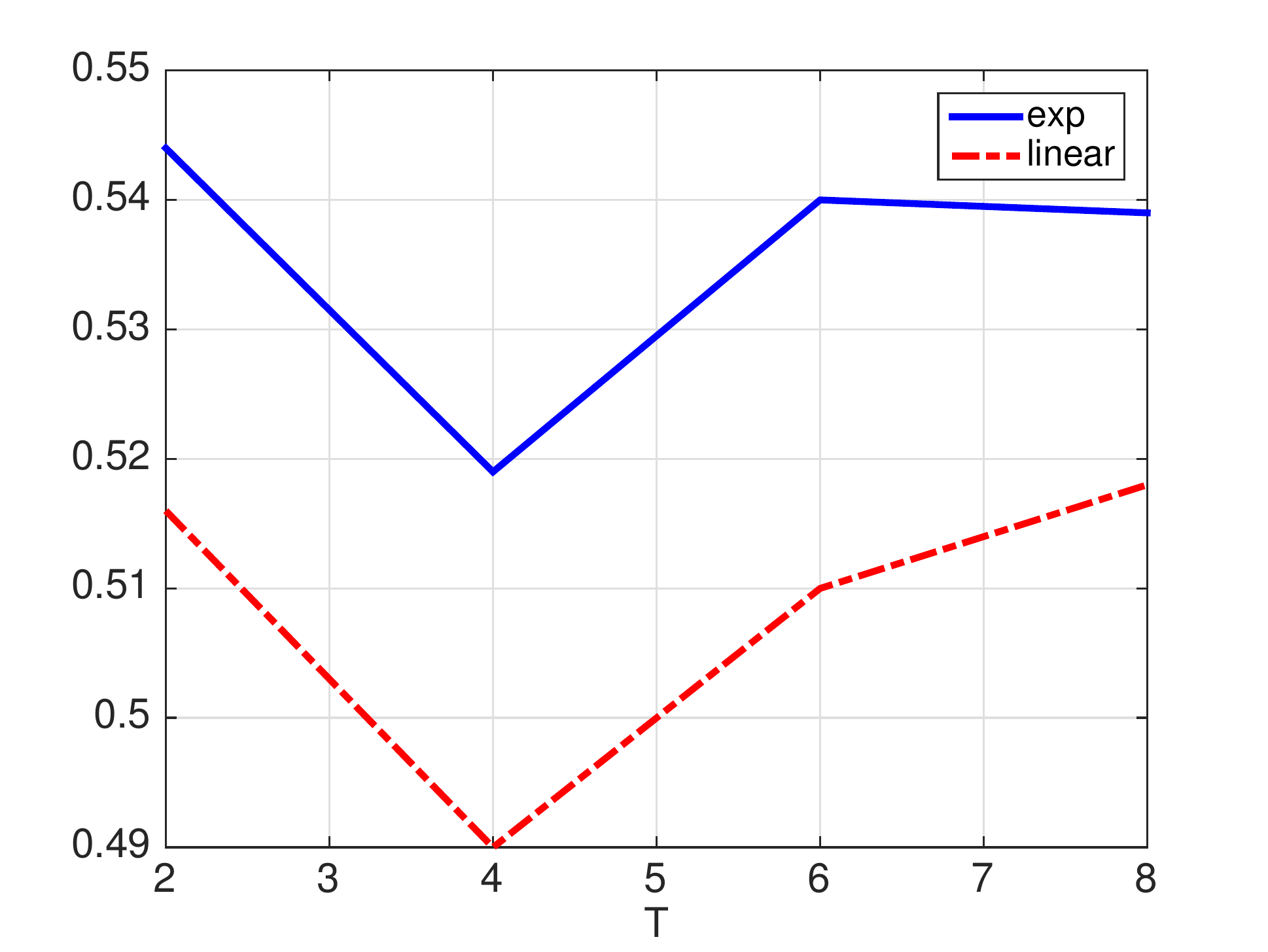}} \hfill
	%\subfloat[][]{    \includegraphics[width = 35mm, height = 30mm]{mlmap-eps-converted-to.pdf}} \hfill
	%\subfloat[][]{     \includegraphics[width = 35mm, height = 30mm]{mlndcg-eps-converted-to.pdf}} \\
	\subfloat[][]{    \includegraphics[width = 35mm, height = 30mm]{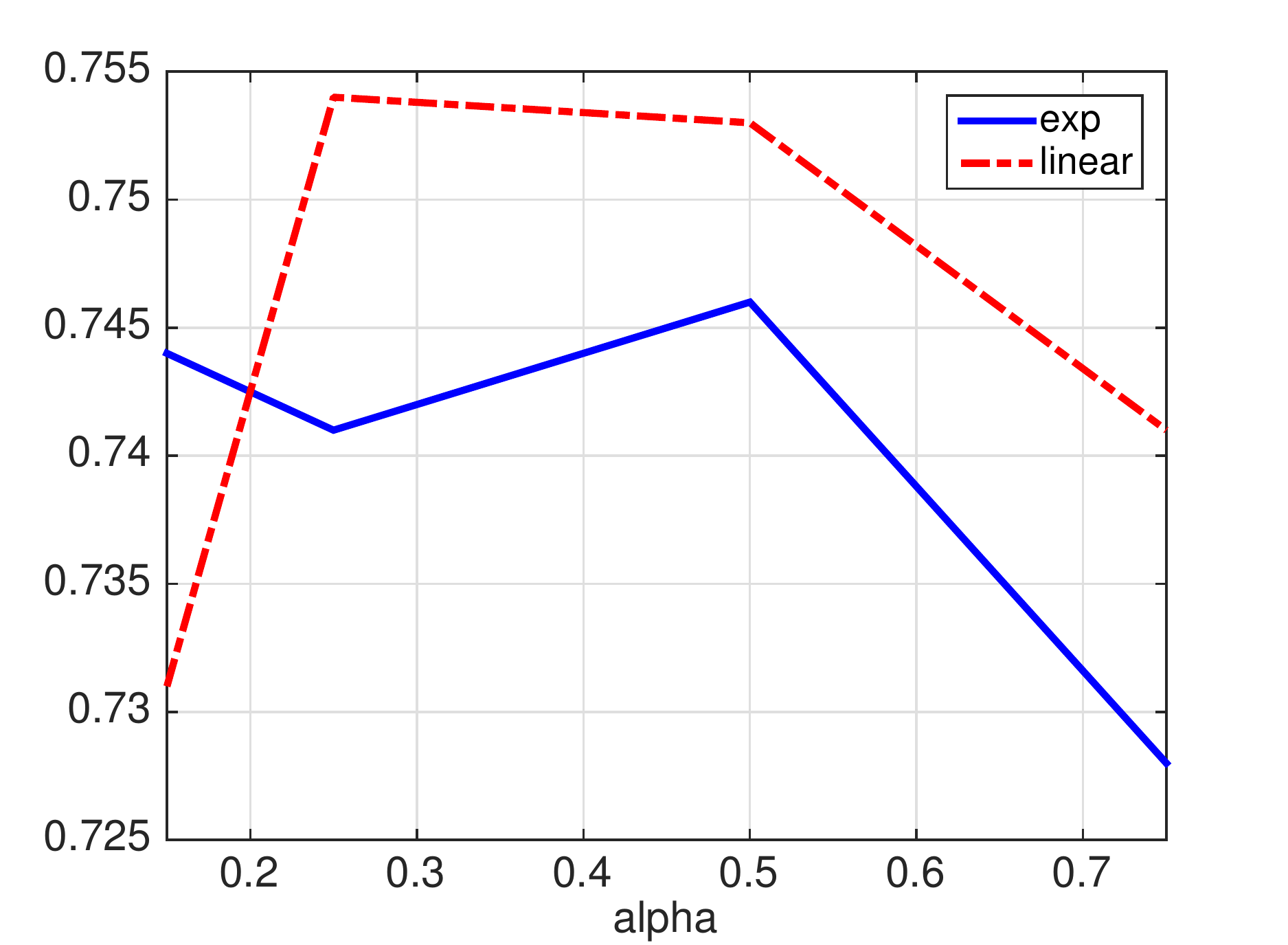}} \hfill
	\subfloat[][]{    \includegraphics[width = 35mm, height = 30mm]{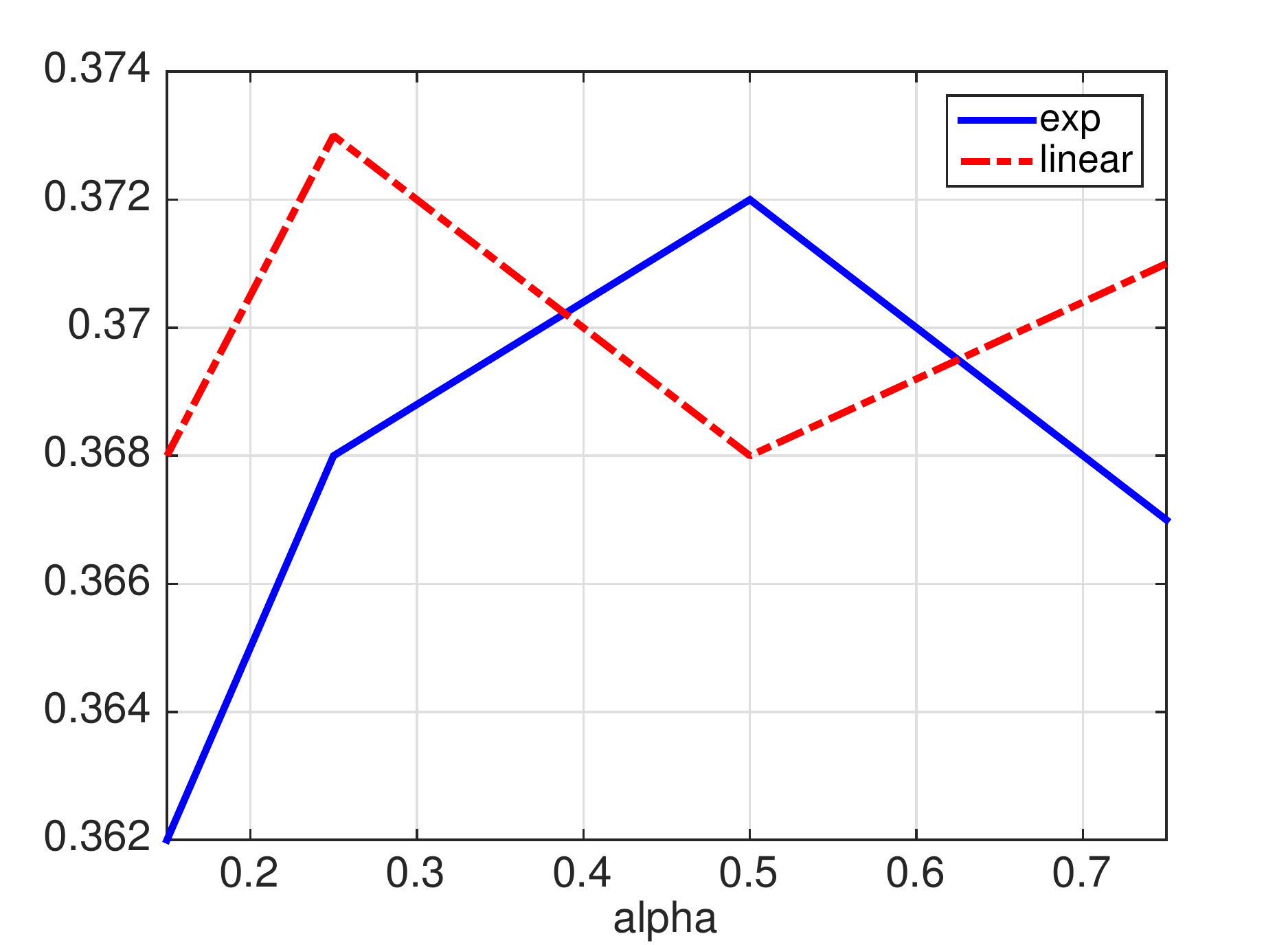}} \hfill
	%\subfloat[][]{     \includegraphics[width = 35mm, height = 30mm]{ftmap2-eps-converted-to.pdf}} \hfill
	%\subfloat[][]{     \includegraphics[width = 35mm, height = 30mm]{ftndcg2-eps-converted-to.pdf}}
	\label{fig:perfT}
\end{figure*}

Next, we show that the walk length is a (hyper)parameter of our method, and should be tuned during cross validation. Figure \ref{fig:perfT} (a)-(b) shows the effect of varying $T$ for the ML-1M dataset. We see that bigger is not necessarily better when it comes to $T$ but for higher orders $(T>2)$, we usually get superior performance in terms of the evaluation metrics.

We also studied the effect of varying $\alpha$ on the FilmTrust dataset, where we fixed $T=4$. Figure \ref{fig:perfT} (c)-(d) again shows that different values of $\alpha$ yield different results, though they are fairly uniform.  

\begin{figure}[!bp]
	\centering
	\includegraphics[width = 50mm, height = 50mm]{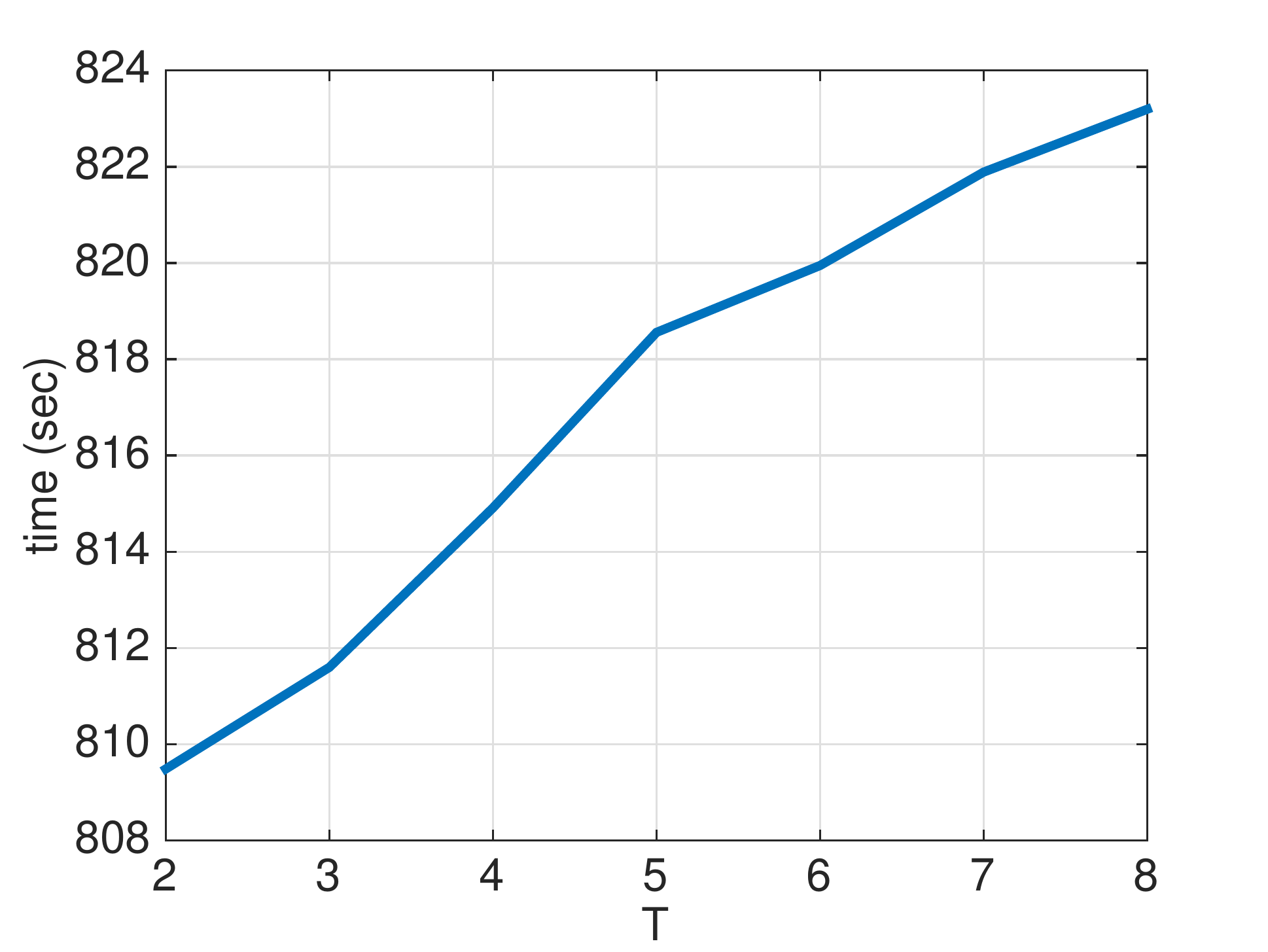}
	\caption{Time taken as a function of walk length}
	\label{fig:T}
\end{figure}

We now show the effect of the parameter $T$ (Figs. \ref{fig:perfT}) and  \ref{fig:T}, the length of walks considered, in \algname, as well as the weighting parameter $\alpha$ (Fig. \ref{fig:perfT}) when the side information graphs are present. We also consider both the $\exp(.)$ and $linear$ functions for $g_1(.), g_2(.)$

Figure \ref{fig:T} shows the effect of varying the walk length $T$ on the runtime of the algorithm, on the ML-1M dataset. %We scale the data so that the time taken for $T=1$ is set to 1 second. 
As expected, the time increases with T. Importantly, note that the time increases nearly linearly with $T$ (with a very small slope), as indicated by the complexity argument in Section \ref{sec:complexity}. However, it also has to be noted that the time difference is not much: approximately 15 seconds extra for the method when $T=8$, compared to $T=2$ seems to suggest that the method is scalable. From a theoretical perspective, even though it may not be useful to take very long walks ($T>20$) \footnote{for most real world graphs, very large T would mean a dense matrix to factorize},  it is impressive to note that long walk lengths do not hamper computational performance.

\begin{table*}[!htbp]
	\captionsetup{justification=centering}
	\caption{Comparison of top two performing algorithms using Top-N evaluation metrics at various percentage of training data}
	\label{tab:ressparse}
	\begin{center}
		\begin{tabular}{| l | l | c | c | c | c | c | c | c | c |}
			\hline 
			% \abovespace\belowspace
			{\textbf{$\%$ data used}} &\textbf{Algorithm} &\multicolumn{2}{|c|}{\textbf{Precision}} &\multicolumn{2}{|c|}{\textbf{Recall}} &\multicolumn{2}{|c|}{\textbf{MAP}} &\multicolumn{2}{|c|}{\textbf{NDCG}}\\
			\hline
			& & $@5$ & $@10$ & $@5$ & $@10$ & $@5$ & $@10$ & $@5$ & $@10$ \\
			\hline
			25 & \algname	&\textbf{0.342} &\textbf{0.307} &\textbf{0.514} & \textbf{0.630} &\textbf{0.461} &\textbf{0.500} &\textbf{0.714} &\textbf{0.721}	 	\\
			& WMF &0.329 &0.296 &0.499 &0.608 &0.452 &0.491 &0.706 &0.709\\
			\hline 
			50 & \algname &\textbf{0.361} &\textbf{0.329} &\textbf{0.536} &\textbf{0.665} &\textbf{0.466} &\textbf{0.507} &\textbf{0.718} & \textbf{0.734}		 	\\
			& WMF &0.325 &0.296 &0.498 &0.613 &0.443 &0.487 &0.699 &0.705\\
			\hline
			75 & \algname	&\textbf{0.372} &\textbf{0.329} &\textbf{0.554} &\textbf{0.672} &\textbf{0.502} &\textbf{0.538} &\textbf{0.740} &\textbf{0.746}	 	\\
			& WMF &0.324 &0.293 &0.497 &0.606 &0.443 &0.484 &0.697 &0.701\\
			\hline
			100 &\algname	& \textbf{0.370}	& \textbf{0.331}	&\textbf{0.544}	&\textbf{0.662}	& \textbf{0.499}	& 0.529	& \textbf{0.744}	& \textbf{0.749} \\
			& WMF	 & 0.357 & 0.329 &0.532 & 0.649 &0.475 & \textbf{0.535} &0.653 & 0.744	\\
			\hline
		\end{tabular}
	\end{center}
\end{table*}

\subsection{Effect of Sparse Data}
The experiments were performed by randomly subsampling the training data but the best hyperparameter search and evaluation was performed on the complete validation and test sets respectively. Table \ref{tab:ressparse} shows a marginal degradation of performance on the ML1M dataset even when we hide 50-75$\%$ of the data. MGCNN, which is similar in spirit also performs similarly, but methods such as WMF, which do not account for higher order interactions show a significant drop in performance. 

\begin{figure}[!bp]
	\captionsetup{justification=centering}
	\centering
	\includegraphics[width = 50mm, height = 40mm]{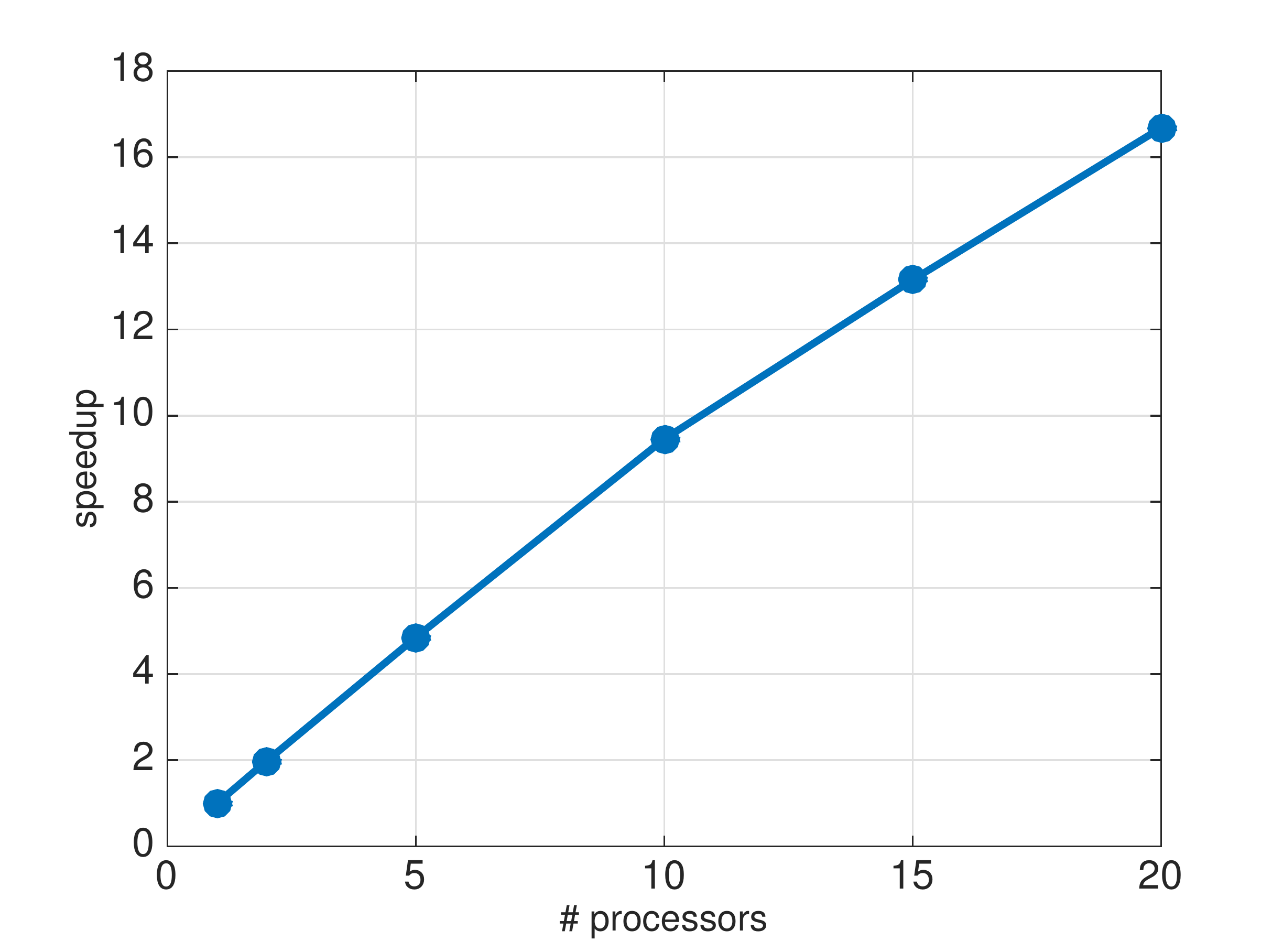}
	\caption{Speedup obtained as a function of $\#$ processors}
	\label{fig:speedup}
\end{figure}

\subsection{Speedup via Parallelization}
For a given number of processors $N$, we define the speedup to be
\[
speedup(N) = \frac{\text{Time taken with 1 processor}}{\text{Time taken with N processors}}
\]
Figure \ref{fig:speedup} shows that we obtain near-linear speedups via the parallel procedure in Algorithm \ref{alg:full_algo}. We used the ML-1M dataset here and varied the number of processors over which we parallelized the updates. When the number of processors is increased to about 20, we see that the corresponding speedup is nearly 17.

\section{Discussion and Future Work} \label{sec:fw}

Our motivation behind transforming the original rating matrix into a Transition Probability Matrix is to find higher order information from the graph. $A^l$ will encode the $l^{th}$ order information. Our aim is to assimilate as much (non-redundant) higher order information as possible. Another key point is that our approach is highly general, and by setting $T=1$ we can obtain similar performance to standard matrix factorization.%, but the normalization does not affect either the sparsity or performance). 

%Let $u_1, v_1$ be the principal left and right eigenvectors respectively of some irreducible, aperiodic matrix $A$, let  $\lambda_1 (= 1)$ denote its dominant eigenvalue. 
%Now, we know that $\lim_{t\rightarrow \infty} A^t \rightarrow \lambda_1^t u_1 v_1' =u_1 v_1' $ (Perron-Frobenius Theorem). 
%
% \vspace{-3pt}
%\paragraph{P1:} For irreducible aperiodic $A$ \footnote{Note that, in absence of side-information or matrices with special structures, the transformed matrix $A$ can be reducible and periodic, in which case P1 may not be true but can be easily modified to include eigenvectors corresponding to other unit-magnitude eigenvalues.},
%\begin{align*}
%\lim_{T \rightarrow \infty} f_T(A) \rightarrow u_1 v_1'
%\end{align*}

At large $l$, $A^l$ might not capture too much discriminative information as two nodes far apart in the graph are forced to have a similar representation. At low values of $l$, one can miss out on the implicit higher order relationships between nearby nodes of the graph. Hence, it is necessary to be prudent while choosing $l$. 

The development of a fully distributed method remains an open challenge. Given the matrix $A$, we can solve the problem in a parallel fashion on a multicore machine as we have demonstrated in this paper. However, for extremely large datasets, the matrix $A$ might not fit into memory. When $A$ itself is distributed, computing higher order powers can be a challenging task, since Algorithm \ref{alg:colsample} requires access to the full matrix $A$. 

From a theoretical perspective, we would like to address the problem posed in Equation \ref{mf_homf} and obtain convergence guarantees for the low rank matrix factorization for functions of matrix $A$ without explicitly computing the function. As of now, we are unaware of any literature pertaining to matrix factorization of higher orders of $A$, say $A^p$ let alone $f_p(A)$ for general matrices $A$. Another interesting avenue is to compute the minimum number of samples required in $A$ to recover $A^p$ as accurately as possible.

\section{Conclusion}
\label{sec:conc}
In this paper, we presented \algname, a unified method for matrix factorization that can take advantage of both explicit and implicit side information in the data. We developed a scalable method for solving the resulting optimization problem and presented results on several datasets. Our method efficiently includes both existing side information and implicit higher order information. Experiments on datasets with and without side information shows that our method outperforms several baselines, and often also those that rely on deep neural networks to make predictions. 

\bibliographystyle{ieeetr}
\bibliography{ge}

\begin{thebibliography}{10}

\bibitem{koren2009matrix}
Y.~Koren, R.~Bell, C.~Volinsky, {\em et~al.}, ``Matrix factorization techniques
  for recommender systems,'' {\em Computer}, vol.~42, no.~8, pp.~30--37, 2009.

\bibitem{biswas2015inferring}
A.~K. Biswas, M.~Kang, D.-C. Kim, C.~H. Ding, B.~Zhang, X.~Wu, and J.~X. Gao,
  ``Inferring disease associations of the long non-coding rnas through
  non-negative matrix factorization,'' {\em Network Modeling Analysis in Health
  Informatics and Bioinformatics}, vol.~4, no.~1, pp.~1--17, 2015.

\bibitem{haeffele2014structured}
B.~Haeffele, E.~Young, and R.~Vidal, ``Structured low-rank matrix
  factorization: Optimality, algorithm, and applications to image processing,''
  in {\em Proceedings of the 31st International Conference on Machine Learning
  (ICML-14)}, pp.~2007--2015, 2014.

\bibitem{recht2011hogwild}
B.~Recht, C.~Re, S.~Wright, and F.~Niu, ``Hogwild: A lock-free approach to
  parallelizing stochastic gradient descent,'' in {\em Advances in Neural
  Information Processing Systems}, pp.~693--701, 2011.

\bibitem{yu2012scalable}
H.-F. Yu, C.-J. Hsieh, S.~Si, and I.~Dhillon, ``Scalable coordinate descent
  approaches to parallel matrix factorization for recommender systems,'' in
  {\em 2012 IEEE 12th International Conference on Data Mining}, pp.~765--774,
  IEEE, 2012.

\bibitem{candes2009exact}
E.~J. Cand{\`e}s and B.~Recht, ``Exact matrix completion via convex
  optimization,'' {\em Foundations of Computational mathematics}, vol.~9,
  no.~6, pp.~717--772, 2009.

\bibitem{cai2010singular}
J.-F. Cai, E.~J. Cand{\`e}s, and Z.~Shen, ``A singular value thresholding
  algorithm for matrix completion,'' {\em SIAM Journal on Optimization},
  vol.~20, no.~4, pp.~1956--1982, 2010.

\bibitem{bhojanapalli2014tighter}
S.~Bhojanapalli, P.~Jain, and S.~Sanghavi, ``Tighter low-rank approximation via
  sampling the leveraged element,'' in {\em Proceedings of the twenty-sixth
  annual ACM-SIAM symposium on Discrete algorithms}, pp.~902--920, SIAM, 2014.

\bibitem{mnih2008probabilistic}
A.~Mnih and R.~R. Salakhutdinov, ``Probabilistic matrix factorization,'' in
  {\em Advances in neural information processing systems}, pp.~1257--1264,
  2008.

\bibitem{crandall2008feedback}
D.~Crandall, D.~Cosley, D.~Huttenlocher, J.~Kleinberg, and S.~Suri, ``Feedback
  effects between similarity and social influence in online communities,'' in
  {\em Proceedings of the 14th ACM SIGKDD international conference on Knowledge
  discovery and data mining}, pp.~160--168, ACM, 2008.

\bibitem{zhou2012kernelized}
T.~Zhou, H.~Shan, A.~Banerjee, and G.~Sapiro, ``Kernelized probabilistic matrix
  factorization: Exploiting graphs and side information.,'' in {\em SDM},
  vol.~12, pp.~403--414, SIAM, 2012.

\bibitem{jamali2009trustwalker}
M.~Jamali and M.~Ester, ``Trustwalker: a random walk model for combining
  trust-based and item-based recommendation,'' in {\em Proceedings of the 15th
  ACM SIGKDD international conference on Knowledge discovery and data mining},
  pp.~397--406, ACM, 2009.

\bibitem{xu2013speedup}
M.~Xu, R.~Jin, and Z.-H. Zhou, ``Speedup matrix completion with side
  information: Application to multi-label learning,'' in {\em Advances in
  Neural Information Processing Systems}, pp.~2301--2309, 2013.

\bibitem{rao2015collaborative}
N.~Rao, H.-F. Yu, P.~K. Ravikumar, and I.~S. Dhillon, ``Collaborative filtering
  with graph information: Consistency and scalable methods,'' in {\em Advances
  in Neural Information Processing Systems}, pp.~2107--2115, 2015.

\bibitem{Liang2016cofactor}
D.~Liang, J.~Altosaar, L.~Charlin, and D.~M. Blei, ``Factorization meets the
  item embedding: Regularizing matrix factorization with item co-occurrence,''
  in {\em Proceedings of the 10th ACM Conference on Recommender Systems},
  RecSys '16, pp.~59--66, 2016.

\bibitem{mikolov2013distributed}
T.~Mikolov, I.~Sutskever, K.~Chen, G.~S. Corrado, and J.~Dean, ``Distributed
  representations of words and phrases and their compositionality,'' in {\em
  Advances in neural information processing systems}, pp.~3111--3119, 2013.

\bibitem{levy2014neural}
O.~Levy and Y.~Goldberg, ``Neural word embedding as implicit matrix
  factorization,'' in {\em Advances in neural information processing systems},
  pp.~2177--2185, 2014.

\bibitem{perozzi2014deepwalk}
B.~Perozzi, R.~Al-Rfou, and S.~Skiena, ``Deepwalk: Online learning of social
  representations,'' in {\em Proceedings of the 20th ACM SIGKDD international
  conference on Knowledge discovery and data mining}, pp.~701--710, ACM, 2014.

\bibitem{tang2015line}
J.~Tang, M.~Qu, M.~Wang, M.~Zhang, J.~Yan, and Q.~Mei, ``Line: Large-scale
  information network embedding,'' in {\em Proceedings of the 24th
  International Conference on World Wide Web}, pp.~1067--1077, 2015.

\bibitem{tu2016max}
C.~Tu, W.~Zhang, Z.~Liu, and M.~Sun, ``Max-margin deepwalk: discriminative
  learning of network representation,'' in {\em Proceedings of the Twenty-Fifth
  International Joint Conference on Artificial Intelligence (IJCAI 2016)},
  pp.~3889--3895, 2016.

\bibitem{goyal2017graph}
P.~Goyal and E.~Ferrara, ``Graph embedding techniques, applications, and
  performance: A survey,'' {\em arXiv preprint arXiv:1705.02801}, 2017.

\bibitem{yang2015comprehend}
C.~Yang and Z.~Liu, ``Comprehend deepwalk as matrix factorization,'' {\em arXiv
  preprint arXiv:1501.00358}, 2015.

\bibitem{defferrard2016convolutional}
M.~Defferrard, X.~Bresson, and P.~Vandergheynst, ``Convolutional neural
  networks on graphs with fast localized spectral filtering,'' in {\em Advances
  in Neural Information Processing Systems}, pp.~3844--3852, 2016.

\bibitem{mgcnn}
F.~Monti, M.~Bronstein, and X.~Bresson, ``Geometric matrix completion with
  recurrent multi-graph neural networks,'' in {\em Advances in Neural
  Information Processing Systems}, 2017.

\bibitem{jain2013provable}
P.~Jain and I.~S. Dhillon, ``Provable inductive matrix completion,'' {\em arXiv
  preprint arXiv:1306.0626}, 2013.

\bibitem{gimc}
S.~Si, K.-Y. Chiang, C.-J. Hsieh, N.~Rao, and I.~S. Dhillon, ``Goal directed
  inductive matrix completion,'' {\em KDD}, 2016.

\bibitem{gori2007itemrank}
M.~Gori, A.~Pucci, V.~Roma, and I.~Siena, ``Itemrank: A random-walk based
  scoring algorithm for recommender engines.,'' in {\em IJCAI}, vol.~7,
  pp.~2766--2771, 2007.

\bibitem{xie2015edge}
W.~Xie, D.~Bindel, A.~Demers, and J.~Gehrke, ``Edge-weighted personalized
  pagerank: breaking a decade-old performance barrier,'' in {\em Proceedings of
  the 21th ACM SIGKDD International Conference on Knowledge Discovery and Data
  Mining}, pp.~1325--1334, ACM, 2015.

\bibitem{yun2014nomad}
H.~Yun, H.-F. Yu, C.-J. Hsieh, S.~Vishwanathan, and I.~Dhillon, ``Nomad:
  Non-locking, stochastic multi-machine algorithm for asynchronous and
  decentralized matrix completion,'' {\em Proceedings of the VLDB Endowment},
  vol.~7, no.~11, pp.~975--986, 2014.

\bibitem{golbeck2005computing}
J.~A. Golbeck, ``Computing and applying trust in web-based social networks,''
  {\em ACM Digital Library}, 2005.

\bibitem{fouss2007random}
F.~Fouss, A.~Pirotte, J.-M. Renders, and M.~Saerens, ``Random-walk computation
  of similarities between nodes of a graph with application to collaborative
  recommendation,'' {\em IEEE Transactions on knowledge and data engineering},
  vol.~19, no.~3, pp.~355--369, 2007.

\bibitem{abbassi2007recommender}
Z.~Abbassi and V.~S. Mirrokni, ``A recommender system based on local random
  walks and spectral methods,'' in {\em Proceedings of the 9th WebKDD and 1st
  SNA-KDD 2007 workshop on Web mining and social network analysis},
  pp.~102--108, ACM, 2007.

\bibitem{yang2015network}
C.~Yang, Z.~Liu, D.~Zhao, M.~Sun, and E.~Y. Chang, ``Network representation
  learning with rich text information,'' in {\em Proceedings of the 24th
  International Joint Conference on Artificial Intelligence, Buenos Aires,
  Argentina}, pp.~2111--2117, 2015.

\bibitem{thyfriend}
C.~Borgs, J.~Chayes, C.~E. Lee, and D.~Shah, ``Thy friend is my friend:
  Iterative collaborative filtering for sparse matrix estimation,'' in {\em
  Advances in Neural Information Processing Systems}, pp.~4718--4729, 2017.

\bibitem{kipf2016semi}
T.~N. Kipf and M.~Welling, ``Semi-supervised classification with graph
  convolutional networks,'' {\em arXiv preprint arXiv:1609.02907}, 2016.

\bibitem{harper2016movielens}
F.~M. Harper and J.~A. Konstan, ``The movielens datasets: History and
  context,'' {\em ACM Transactions on Interactive Intelligent Systems (TiiS)},
  vol.~5, no.~4, p.~19, 2016.

\bibitem{guo2013novel}
G.~Guo, J.~Zhang, and N.~Yorke-Smith, ``A novel bayesian similarity measure for
  recommender systems,'' in {\em Proceedings of the 23rd International Joint
  Conference on Artificial Intelligence (IJCAI)}, pp.~2619--2625, 2013.

\bibitem{hu2008collaborative}
Y.~Hu, Y.~Koren, and C.~Volinsky, ``Collaborative filtering for implicit
  feedback datasets,'' in {\em Data Mining, 2008. ICDM'08. Eighth IEEE
  International Conference on}, pp.~263--272, Ieee, 2008.

\end{thebibliography}

\newpage
\section{Appendix}

\textbf{A Simple Example}\\
Let us consider a simple ratings matrix, R:
\begin{align}
{R} = \begin{bmatrix}
& 2 ~~~ &4~~~~~~ \\ 
~~~ &- ~~~ &3~~~~~ 
\end{bmatrix}
\end{align}
In this part, we will describe how to construct $A$ from $R$ and $G_c, G_r$ (if given). We will assume that we use the exponential weighting (any non-negative non-decreasing mapping can be used)of the ratings to construct the transition probability matrix $A$.
 $G$ then becomes,
 \begin{align}
 {G} &= \begin{bmatrix}
 & 0 ~~~ &g_2(R)~~~  \\ 
 ~~~ &g_2(R)^T ~~~ &0~~~~~ 
 \end{bmatrix}\\
 &= \begin{bmatrix}
 & 0 ~~~ &0~~~  & \exp(2) ~~~ &\exp(4)~~~  \\ 
  & 0 ~~~ &0~~~  &0~~~ &\exp(3)~~~\\
  & \exp(2) ~~~ &0~~~  &0~~~ &0~~~ \\
  &\exp(4) ~~~& \exp(3)~~~  &0~~~ &0~~~ 
 \end{bmatrix}
 \end{align}
 
 Let us assume that we have no side information for the moment.
 The row normalized version of $G$ i.e. $A$ is given as follows:
  \begin{align}
 {A} = \begin{bmatrix}&0~~~ &0~~~ &\dfrac{exp(2)}{\exp(2) + \exp(4)}  & \dfrac{exp(4)}{\exp(2) + \exp(4)}~~~  \\ 
 ~~~ &0 ~~~ &0~~~ &0 ~~~ &1~~~ \\
  ~~~ &1 ~~~ &0~~~ &0 ~~~ &0~~~ \\
   ~~~ &\dfrac{exp(4)}{\exp(3) + \exp(4)}  ~~~ &\dfrac{exp(3)}{\exp(3) + \exp(4)} ~~~ &0 ~~~ &0~~~ 
 \end{bmatrix}  \label{eq:matA}
 \end{align}
 
 In presence of side information $G_c$ for movies, $G$ becomes,
  \begin{align}
 {G} = \begin{bmatrix}
 & 0 ~~~ &(1-\alpha)g_2(R)~~~  \\ 
 ~~~ &(1-\alpha)g_2(R)^T ~~~ &\alpha G_c~~~~~
 \end{bmatrix}
 \end{align}
 where  
 \begin{align}
 {G_c} = \begin{bmatrix}
 & 0 ~~~ &1~~~  \\ 
 ~~~ &1 ~~~ &0~~~~~ 
 \end{bmatrix}
 \end{align}
 
 Using the same exponential decompostion (or a linear one), we can rewrite A as:
 \begin{align}
 {A} = \begin{bmatrix}&0 &0 &\dfrac{exp(2)}{\exp(2) + \exp(4)}  & \dfrac{exp(4)}{\exp(2) + \exp(4)}~~~  \\ 
&0  &0&0&1~~~ \\
  &\dfrac{(1-\alpha)\exp(2)}{d_1}   &0 &0&\dfrac{\alpha\exp(1)}{d_1}  \\
  &\dfrac{(1-\alpha)\exp(4)}{d_2} &\dfrac{(1-\alpha)\exp(3)}{d_2} &\dfrac{\alpha\exp(1)}{d_2}  &0
 \end{bmatrix}  \label{eq:matA}
 \end{align}
where $d_1 = (1-\alpha) \exp(2) + \alpha \exp(1), d_2 = (1-\alpha) (\exp(4) + \exp(3))  + \alpha \exp(1)$

\subsection{Some observations}

In this section, we will enlist some of the properties we came across for $A$, which we believe should be useful for theoretical guarantees:

\paragraph{P1:} If $\lambda$ is any eigenvalue of $A$, then the eigenvalue of $f_T(A)$ is given as $h(\lambda, T)$, where 
\begin{align*}
h(\lambda, T) = \dfrac{\sum_{t=1}^T \lambda^t}{T} =\begin{cases}  \dfrac{\lambda (1 - \lambda^T)}{(1-\lambda)T} &\text{ if } \lambda < 1\\
1 &\text{ if } \lambda = 1
\end{cases}
\end{align*}

\paragraph{P2:}$h(\lambda, T)$ is a monotonic non-increasing function of $T$ (for $T>1$) and non-decreasing function of $\lambda$ for $\lambda \in (0,1)$.

\begin{figure}[!h]
	\centering
	\includegraphics[width = 60mm, height = 50mm]{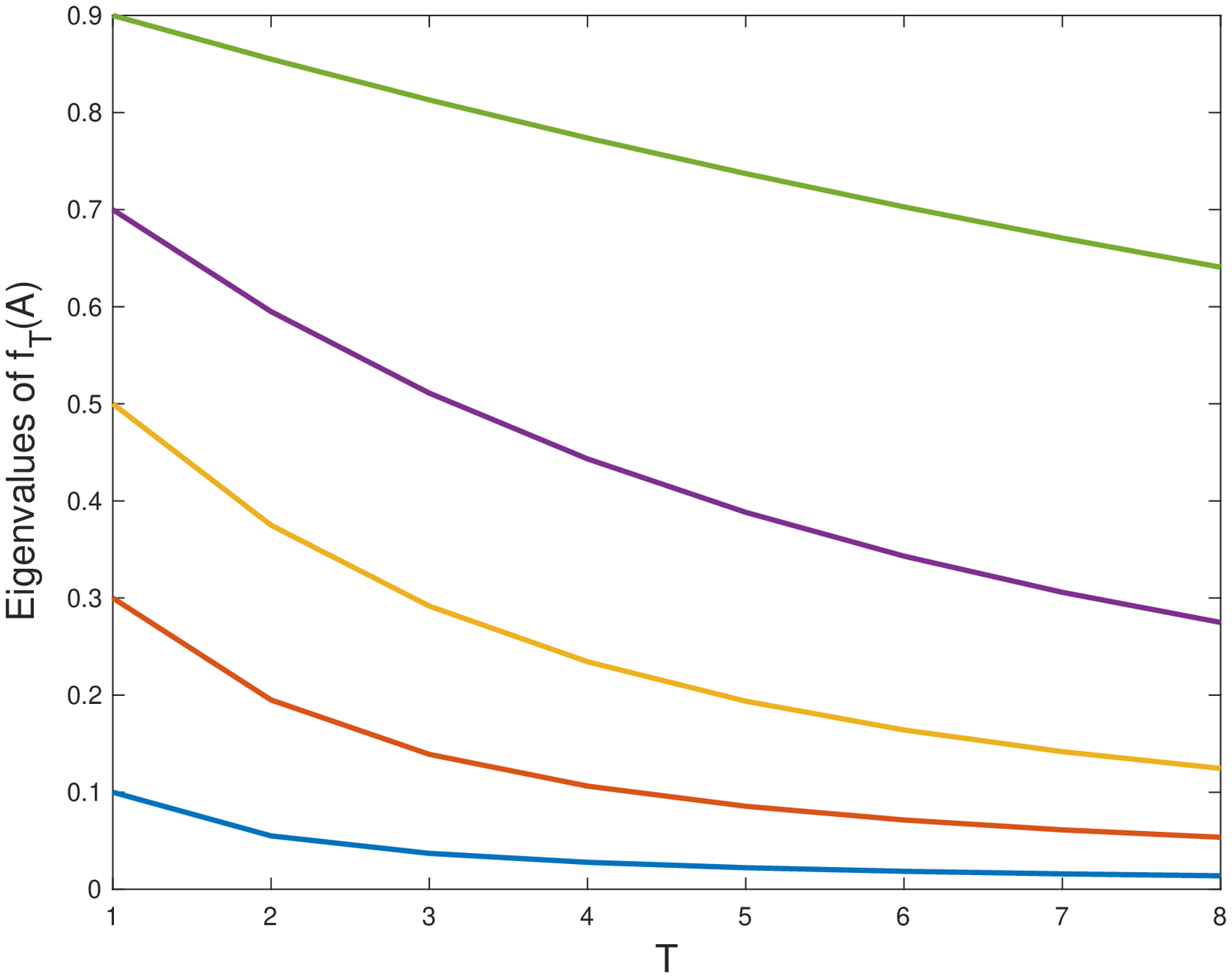}
	\caption{Eigenvalue decay as a function of T (walk length)}
	\label{fig:eigT}
\end{figure}

\noindent
The motivation of using $f_T(A)$ is two-fold: retention of the higher order information and ensuring the existence of low rank factorization. The latter is evident by P2 and Figure  \ref{fig:eigT}.

In Figure \ref{fig:eigT}, we demonstrate that the magnitude of all the eigenvalues  \footnote{Note that these eigenvalues do not correspond to a particular dataset but are representative for the trajectories the actual eigenvalues of $f_T(A)$ will follow.} $( <1)$ decay as we collect higher-order information. This buttresses our claims that $f_T(A)$ for large $T$ can be approximated with a low-rank factorization.

\paragraph{P3:} The trace of $A^k$ is positive for any $k>2$ and so $A$ is not nilpotent.

\noindent
As we allow the random walks to travel back to the same entity (user/movie) from which it began in two or more steps, the probability of returning back to the same entity is non-zero and so the trace of the amtrix is always non-zero for higher powers of $A$.

\paragraph{P4:} Assuming the matrix does not have all-zero rows/columns, $\norm{A} \geq 1$\\
\noindent
Assuming no rows/columns with all zero entries, there has to be atleast one column with a sum greater than 1, since all rows sum to 1. Since the matrix $A$ is not doubly stochastic, its norm is greater than 1.

%\paragraph{P5:} Let $U_1, V_1$ be the low rank matrix decomposition of $A= f_1(A)$, such that $\norm{A - U_1 V_1^T} \leq \epsilon$, then $\norm{f_T(A) - f_T(U_1 V_1^T)} \leq \epsilon O(\norm{A}^{T+1})$

%\textbf{Proof:}
%We can show this by induction:

\end{document}